\documentclass[10pt,conference]{IEEEtran}
\IEEEoverridecommandlockouts
\usepackage[table,xcdraw]{xcolor}

\usepackage{cite}
\usepackage{amsmath,amssymb,amsfonts}
\usepackage{algorithmic}
\usepackage{graphicx}
\usepackage{textcomp}
\usepackage[absolute]{textpos}

\usepackage{enumitem}
\usepackage{multirow}
\usepackage{lipsum}
\usepackage{listings}
\usepackage{subfigure}
\usepackage{graphicx}
\usepackage{pdfpages}
\usepackage{verbatim}
\usepackage{pifont}
\usepackage{etoolbox}
\usepackage{textcomp}
\usepackage{ulem}
\usepackage{hyperref}
\usepackage{url}
\usepackage{caption}
\usepackage{booktabs} 
\usepackage{siunitx} 
\usepackage{etoolbox}
\usepackage[switch]{lineno} 

\usepackage[linesnumbered,titlenumbered,ruled,vlined,resetcount,algosection]{algorithm2e}
\usepackage{xcolor}
\usepackage{color}
\def\BibTeX{{\rm B\kern-.05em{\sc i\kern-.025em b}\kern-.08em
    T\kern-.1667em\lower.7ex\hbox{E}\kern-.125emX}}

\renewcommand{\footnoterule}{%
  \kern -3pt
  \hrule
  \kern 2pt
}

\usepackage{caption}
\def\BibTeX{{\rm B\kern-.05em{\sc i\kern-.025em b}\kern-.08em
    T\kern-.1667em\lower.7ex\hbox{E}\kern-.125emX}}
\newsavebox{\mybox}

\begin{document}

\Urlmuskip=0mu plus 1mu



\author{\IEEEauthorblockN{Yujia Zhai,\IEEEauthorrefmark{1}\IEEEauthorrefmark{5}
Chengquan Jiang,\IEEEauthorrefmark{2}\IEEEauthorrefmark{5}
Leyuan Wang,\IEEEauthorrefmark{2}
Xiaoying Jia,\IEEEauthorrefmark{2}
Shang Zhang,\IEEEauthorrefmark{3} \\
Zizhong Chen,\IEEEauthorrefmark{1}
Xin Liu,\IEEEauthorrefmark{2}\IEEEauthorrefmark{4}
Yibo Zhu\IEEEauthorrefmark{2}
}
\IEEEauthorblockA{\IEEEauthorrefmark{1}University of California, Riverside}
\IEEEauthorblockA{\IEEEauthorrefmark{2}ByteDance Ltd.}
\IEEEauthorblockA{\IEEEauthorrefmark{3}NVIDIA Corporation}
\IEEEauthorblockA{\IEEEauthorrefmark{4}Correspondence to liuxin.ai@bytedance.com}
\IEEEauthorblockA{\IEEEauthorrefmark{5}These authors contributed equally to this work.}
}


\title{ByteTransformer: A High-Performance Transformer Boosted for Variable-Length Inputs
\thanks{We have made ByteTransformer open-source and available at a public GitHub repository: \href{https://github.com/bytedance/ByteTransformer}{\textcolor{blue}{https://github.com/bytedance/ByteTransformer}}.}
}


\maketitle
\thispagestyle{plain}
\pagestyle{plain}


\begin{abstract}

Transformers have become keystone models in natural language processing over the past decade. They have achieved great popularity in deep learning applications, but the increasing sizes of the parameter spaces required by transformer models generate a commensurate need to accelerate performance. Natural language processing problems are also routinely faced with variable-length sequences, as word counts commonly vary among sentences. Existing deep learning frameworks pad variable-length sequences to a maximal length, which adds significant memory and computational overhead. In this paper, we present ByteTransformer, a high-performance transformer boosted for variable-length inputs. We propose a padding-free algorithm that liberates the entire transformer from redundant computations on zero padded tokens. In addition to algorithmic-level optimization, we provide architecture-aware optimizations for transformer functional modules, especially the performance-critical algorithm Multi-Head Attention (MHA). Experimental results on an NVIDIA A100 GPU with variable-length sequence inputs validate that our fused MHA outperforms PyTorch by 6.13x. The end-to-end performance of ByteTransformer for a forward BERT transformer surpasses state-of-the-art transformer frameworks, such as PyTorch JIT, TensorFlow XLA, Tencent TurboTransformer, Microsoft DeepSpeed-Inference and NVIDIA FasterTransformer, by 87\%, 131\%, 138\%, 74\% and 55\%, respectively. We also demonstrate the general applicability of our optimization methods to other BERT-like models, including ALBERT, DistilBERT, and DeBERTa.


\end{abstract}


\section{Introduction}

The transformer model \cite{vaswani2017attention}is a proven effective architecture widely used in a variety of deep learning (DL) applications, such as language modeling\cite{devlin2018bert, yang2019xlnet}, neural machine translation\cite{edunov2018understanding, vaswani2017attention} and recommendation systems\cite{chen2019behavior, sun2019bert4rec}. The last decade has witnessed rapid developments in natural language processing (NLP) pre-training models based on the transformer model, such as Seq2seq\cite{vaswani2017attention}, GPT-2\cite{radford2019language} and XLNET\cite{yang2019xlnet}, which have also greatly accelerated the progress of NLP. Of all the pre-training models based on transformers, Bidirectional Encoder Representations from Transformers (BERT), proposed in 2018 \cite{devlin2018bert}, is arguably the most seminal, inspiring a series of subsequent works and outperforming reference models on a dozen NLP tasks at the time of creation.

BERT-like models consume increasingly larger parameter space and correspondingly more computational resources. When BERT was discovered, a large model required 340 million parameters \cite{zeng2022boosting}, but currently a full GPT-3 model requires 170 billion parameters \cite{brown2020language}. The base BERT model requires 6.9 billion floating-point operations to inference a 40-word sentence, and this number increases to 20 billion when translating a 20-word sentence using a base Seq2Seq model \cite{fang2021turbotransformers}. The size of the parameter space and the computational demands increase the cost of the training and inference for BERT-like models, which requires the attention of the DL community in order to accelerate these models.

To exploit hardware efficiency, DL frameworks adopt a batching strategy, where multiple batches are executed concurrently. Since batched execution requires task shapes in different batches to be identical, DL frameworks presume fixed-length inputs when designing the software \cite{rajbhandari2020zero, rasley2020deepspeed, shoeybi2019megatron, wang2021lightseq2}. However, this assumption cannot always hold, because transformer models are often faced with variable-length input problems \cite{zeng2022boosting, fang2021turbotransformers}. In order to deploy models with variable-length inputs directly to conventional frameworks that support only fixed-length models, a straightforward solution is to pad all sequences with zeros to the maximal sequence length. However, this immediately brings in redundant computations on wasted padded tokens. These padded zeros also introduce significant memory overhead that can hinder a large transformer model from being efficiently deployed.

Existing popular DL frameworks, such as Google TensorFlow with XLA \cite{abadi2016tensorflow, tf-xla}, Meta PyTorch with JIT \cite{paszke2019pytorch}, and OctoML TVM \cite{cho2014properties}, leverage the domain-specific just-in-time compilation technique to boost performance. Another widely-adopted strategy to generate low-level performance optimization is delicate manual tuning: NVIDIA TensorRT \cite{nv-trt}, a DL runtime, falls into this category. Yet all of these frameworks require the input sequence lengths to be identical to exploit the speedup of batch processing. To lift the restriction on fixed sequence lengths, Tencent \cite{fang2021turbotransformers} and Baidu \cite{zeng2022boosting} provide explicit support for models with variable sequence lengths. They group sequences with similar lengths before launching batched kernels to minimize the padding overhead. However, this proactive grouping approach still introduces irremovable padding overhead when grouping and padding sequences with similar yet different lengths.

In contrast to training processes that can be computed offline, the inference stage of a serving system must be processed online with low latency, which imposes high performance requirements on DL frameworks. A highly efficient DL inference framework for NLP models requires delicate kernel-level optimizations and explicit end-to-end designs to avoid wasted computations on zero tokens when handling variable-length inputs. However, existing DL frameworks do not meet these expectations. In order to remedy this deficit, we present ByteTransformer, a highly efficient transformer framework optimized for variable-length inputs in NLP problems. We not only design an algorithm that frees the entire transformer of padding when dealing with variable-length sequences, but also provide a set of hand-tuned fused GPU kernels to minimize the cost of accessing GPU global memory. More specifically, our contributions include:

\begin{itemize}
\item We design and develop ByteTransformer, a high-performance GPU-accelerated transformer optimized for variable-length inputs. ByteTransformer has been deployed to serve world-class applications including TikTok and Douyin of ByteDance.
\item We propose a padding-free algorithm that packs the input tensor with variable-length sequences and calculates the positioning offset vector for all transformer operations to index, which keeps the whole transformer pipeline free from padding and calculations on zero tokens.
\item We propose a fused Multi-Head Attention (MHA) to alleviate the memory overhead of the intermediate matrix, which is quadratic to the sequence length, in MHA without introducing redundant calculations due to padding for variable-length inputs. Part of our fused MHA has been deployed in the production code base of NVIDIA CUTLASS.
\item We hand-tune the memory footprints of layer normalization, adding bias and activation to squeeze the final performance of the system.
\item We benchmark the performance of ByteTransformer on an NVIDIA A100 GPU for forward pass of BERT-like transformers, including BERT, ALBERT, DistilBERT, and DeBERTa. Experimental results demonstrate our fused MHA outperforms standard PyTorch attention by 6.13X. Regarding the end-to-end performance of standard BERT transformer, ByteTransformer surpasses PyTorch, TensorFlow, Tencent TurboTransformer, Microsoft DeepSpeed and NVIDIA FasterTransformer by 87\%, 131\%, 138\%, 74\%, and 55\%, respectively.
\end{itemize}

The rest of the paper is organized as follows: we introduce background and related works in Section \ref{sec:background}, and then detail our systematic optimization approach in Section \ref{sec:design_and_optimizations}. Evaluation results are given in Section \ref{sec:results}. We conclude our paper and present future work in Section \ref{sec:conclusion}.

\section{Background and Related Works} \label{sec:background}

We provide an overview of the transformer model, including its encoder-decoder architecture and multi-head attention layer. We also survey related works on DL framework acceleration.

\subsection{The transformer architecture}

\begin{figure}[ht] \centering
\scalebox{0.8}{
\includegraphics[width=0.48\textwidth]{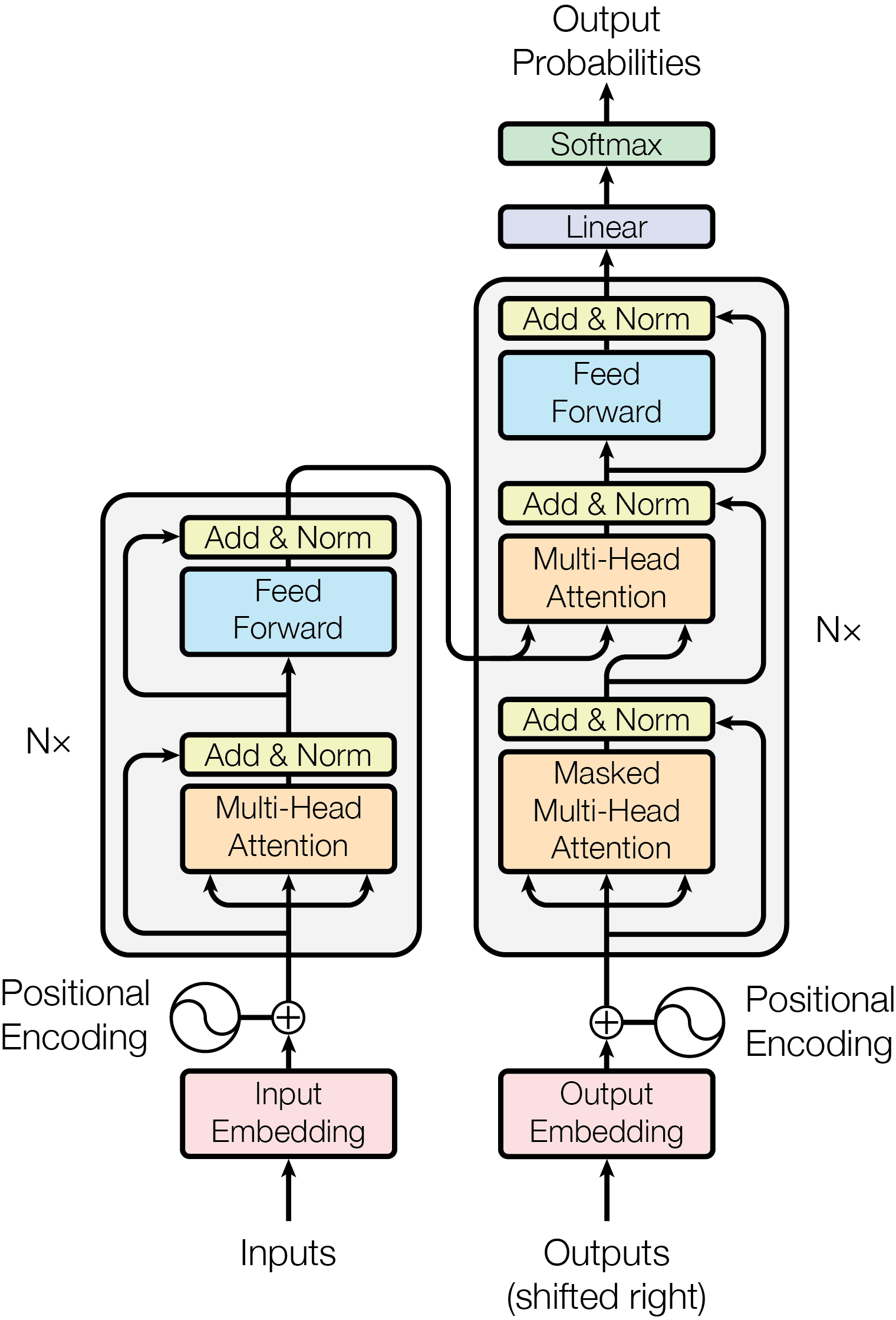}
}
\caption{The transformer architecture. \cite{vaswani2017attention}}
\label{fig:attn-is-all-you-need}
\end{figure}

Figure \ref{fig:attn-is-all-you-need} shows the encoder-decoder model architecture of the transformer. It consists of stacks of multiple encoder and decoder layers. In an encoder layer, there is a multi-head attention layer followed by a feed-forward network (FFN) layer. A layer normalization (layernorm) operation is applied after both MHA and FFN. In a decoder layer, there are two sets of consecutive MHA layers and one FFN layer, and each operation is normalized with a layernorm. The FFN is used to improve the capacity of the model. In practice, FFN is implemented by multiplying the tensor by a larger scaled tensor using GEMM. Here we skip the embedding descriptions in the figure, and refer an interested reader to \cite{vaswani2017attention} for details. Although we show both encoder and decoder modules for this transformer, a BERT transformer model only contains the encoder section \cite{devlin2018bert}. In this paper, we present optimizations for BERT-like transformer models, which can be extended to other transformers containing decoder sections.

Self-attention is a key module of the transformer architecture. Conceptually, self-attention computes the significance of each position of the input sequence, with the information from other positions considered. A self-attention receives three input tensors: query (Q), key (K), and value (V). Self-attention can be split into multiple heads. The Q and K tensors are first multiplied ($1^{st}$ GEMM) to compute the dot product of the query against all keys. This dot product is then scaled by the hidden dimension $d_k$ and passed through a softmax function to calculate the weights corresponding to the value tensor. Each head of the output tensor is concatenated before going through another linear layer by multiplying against tensor V ($2^{nd}$ GEMM). Expressing self-attention as a mathematical formula, we have:
\begin{equation}
    Attention(Q,K,V) = softmax(\frac{QK^T}{\sqrt{d_k}})\times V \\
\end{equation}

\noindent Whereas the formula of multi-head attention is: $Multihead(Q,K,V) = Concat(head_i,...,head_h)$, here $head_i =Attention(Q_i,K_i,V_i)$.

\subsection{Related works on DL acceleration}

Performance is a crucial aspect in the real-world deployment of software systems, attracting significant attention across various applications \cite{zhai2021ft,zhao2020algorithm,zhai2022accelerating}, including DL frameworks. The conventional DL frameworks, such as PyTorch, TensorFlow, TVM, and TensorRT are designed explicitly for fixed-length input tensors. When dealing with NLP problems with variable-length input, all sequences are padded to the maximal length, which leads to significant wasted calculations on zero tokens. A few DL frameworks, such as Tencent TurboTransformer \cite{fang2021turbotransformers} and NVIDIA FasterTransformer \cite{nv-ft}, employ explicit designs for variable-length inputs. TurboTransformer designs run-time algorithms to group and pad sequences with similar lengths to minimize the padding overhead. TurboTransformer also uses a run-time memory scheduling strategy to improve end-to-end performance. Kernel-level optimizations are of the same significance as algorithmic optimizations. NVIDIA's FasterTransformer uses vendor-specific libraries such as TensorRT and cuBLAS \cite{nv-cublas} as its back-end, which provide optimized implementations of various operations at the kernel level.

Other end-to-end DL frameworks have also presented optimizations for BERT-like transformers, such as E.T. \cite{chen2021re} and DeepSpeed-Inference \cite{aminabadi2022deepspeed}. E.T. introduces a novel MHA architecture for NVIDIA Volta GPUs and includes pruning designs for end-to-end transformer models. In contrast, ByteTransformer targets unpruned models and is optimized for NVIDIA Ampere GPUs. DeepSpeed-Inference is optimized for large distributed models on multiple GPUs, while ByteTransformer currently focuses on lighter single-GPU models.


In addition to end-to-end performance acceleration, the research community has also made focused efforts to improve a key algorithm of the transformer, multi-head attention. PyTorch provides a standard implementation of MHA \cite{torch-mha}. NVIDIA TensorRT utilizes a fused MHA for short sequences with lengths up to 512, as described in \cite{nv-mha}. To handle longer sequences, FlashAttention was proposed by Stanford researchers in \cite{dao2022flashattention}. FlashAttention assigns the workload of a whole attention unit to a single threadblock (CTA). However, this approach can result in underutilization on wide GPUs when there are not enough attention units assigned. Our fused MHA, on the other hand, provides high performance for both short and long sequences for variable-length inputs without leading to performance degradation in small-batch scenarios.


\begin{table}[ht] \centering
\caption{Summarizing  state-of-the-art transformers.}
\begin{tabular}{l
    S[table-format=3] 
    S[table-format=3] 
    S[table-format=5] 
    S[table-format=3] 
    }
\toprule
                   & {variable-len}       & {kernel}       & {fused}  & {kernel} \\ 
                     & {support}       & {tuning}       & {MHA} & {fusion}\\ 
\midrule
Tensorflow XLA              & {no}          & {yes}              & {no}    & no      \\ 
PyTorch JIT              & {no}          & {yes}             & {no}    & {no}     \\ 
FasterTransformer              & {yes}         & {yes}             & {$\le512$}      & {no}   \\ 
TurboTransformer             & {yes}         & {yes}             & {no}      & {partially}  \\ 
\textbf{ByteTransformer}             & \textbf{yes}         & \textbf{yes}             & \textbf{yes}      & \textbf{yes}  \\ 
\bottomrule
\end{tabular}
\label{tab:transformer-survey}
\end{table}

Table \ref{tab:transformer-survey} surveys state-of-the-art transformers. TensorFlow and PyTorch provide tuned kernels but require padding for variable-length inputs. NVIDIA FasterTransformer and Tencent TurboTransformer, although providing support for variable-length inputs, do not perform comprehensive kernel fusion or explicit optimization for the hot-spot algorithm MHA for any length of sequence. In addition, TurboTransformer only optimizes part of the fusible operations in the transformer model, such as layernorm and activation, namely 'partial kernel fusion' in the table. Our ByteTransformer, in contrast, starting with a systemic profiling to locate bottleneck algorithms, precisely tunes a series of kernels including the key algorithm MHA. We also propose a padding-free algorithm which completely removes redundant calculations for variable-length inputs from the entire transformer.

\section{Designs and Optimizations}
\label{sec:design_and_optimizations}

In this section, we present our algorithmic and kernel-level optimizations to improve the end-to-end performance of BERT transformer under variable-length inputs.

\subsection{Math expression of BERT transformer encoder}

\begin{figure}[ht] \centering
\includegraphics[width=0.48\textwidth]{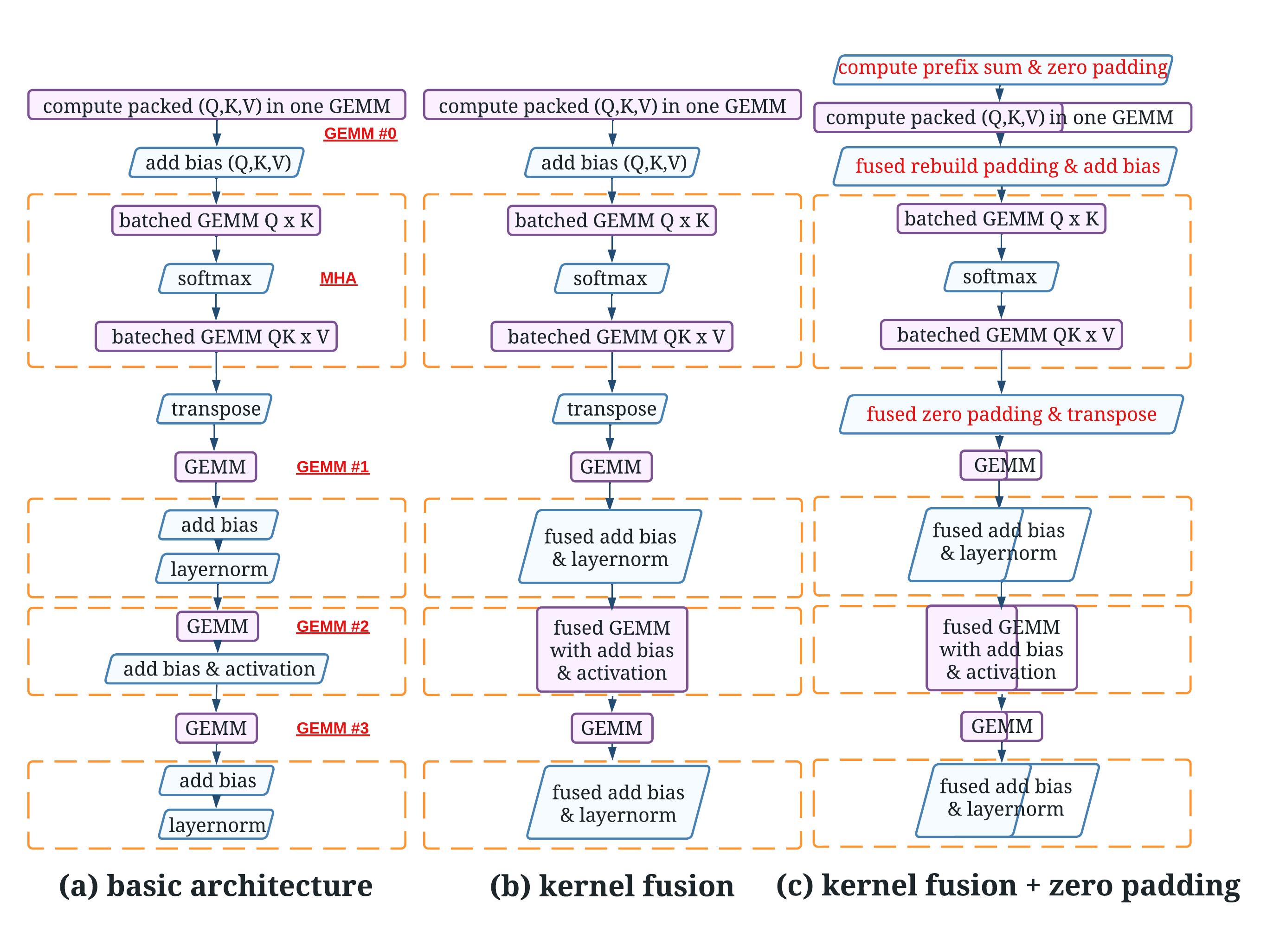}
\caption{BERT transformer architecture and optimizations.}
\label{fig:arch}
\end{figure}

Figure \ref{fig:arch}(a) illustrates the architecture of the transformer encoder. The input tensor is first processed through the BERT pipeline, where it is multiplied by a built-in attribute matrix to perform Q, K, and V positioning encoding. This operation can be implemented using three separate GEMM operations or in batch mode. Realizing that the corresponding attribute matrices to Q, K, and V are all the same shape (hidden\_dim x hidden\_dim), we pack them to continuous memory space and launch a single batched GEMM kernel that calculates Q, K, and V to reduce the kernel launch overhead at runtime. Bias matrices for Q, K, and V are then added to the encoded tensor, which is passed through the self-attention module. In addition to the multi-head attention module, the BERT transformer encoder includes projection, feed forward network, and layer normalization. The encoder pipeline can be represented as a series of mathematical operations, including six GEMMs (shown in light purple) and other memory-bound operations (shown in light blue).

\subsection{Profiling for single-layer standard BERT transformer}

We implement the pipeline of Figure \ref{fig:arch} (a) by calling cuBLAS and profile its single-layer performance on an NVIDIA A100 GPU. We adopt the standard BERT transformer configuration (batch size: 16, head number: 12, head size: 64) and profile for two different sequence lengths: 256 and 1024.

\begin{figure}[ht] \centering
\subfigure[{Sequence lengths 256}]
{
\includegraphics[width=0.225\textwidth]{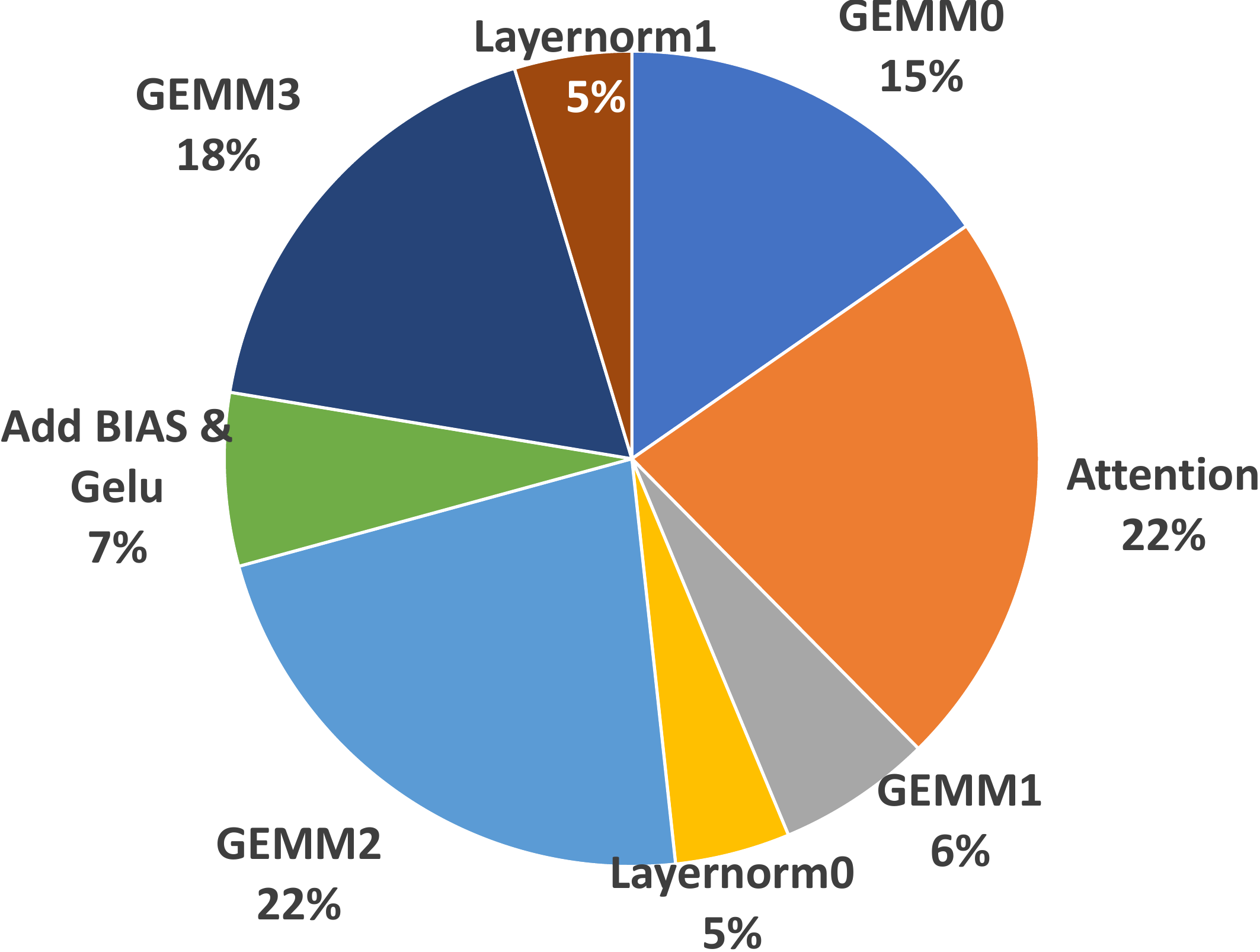}
}
\hspace{-3mm}
\subfigure[Sequence lengths 1024]
{
\includegraphics[width=0.225\textwidth]{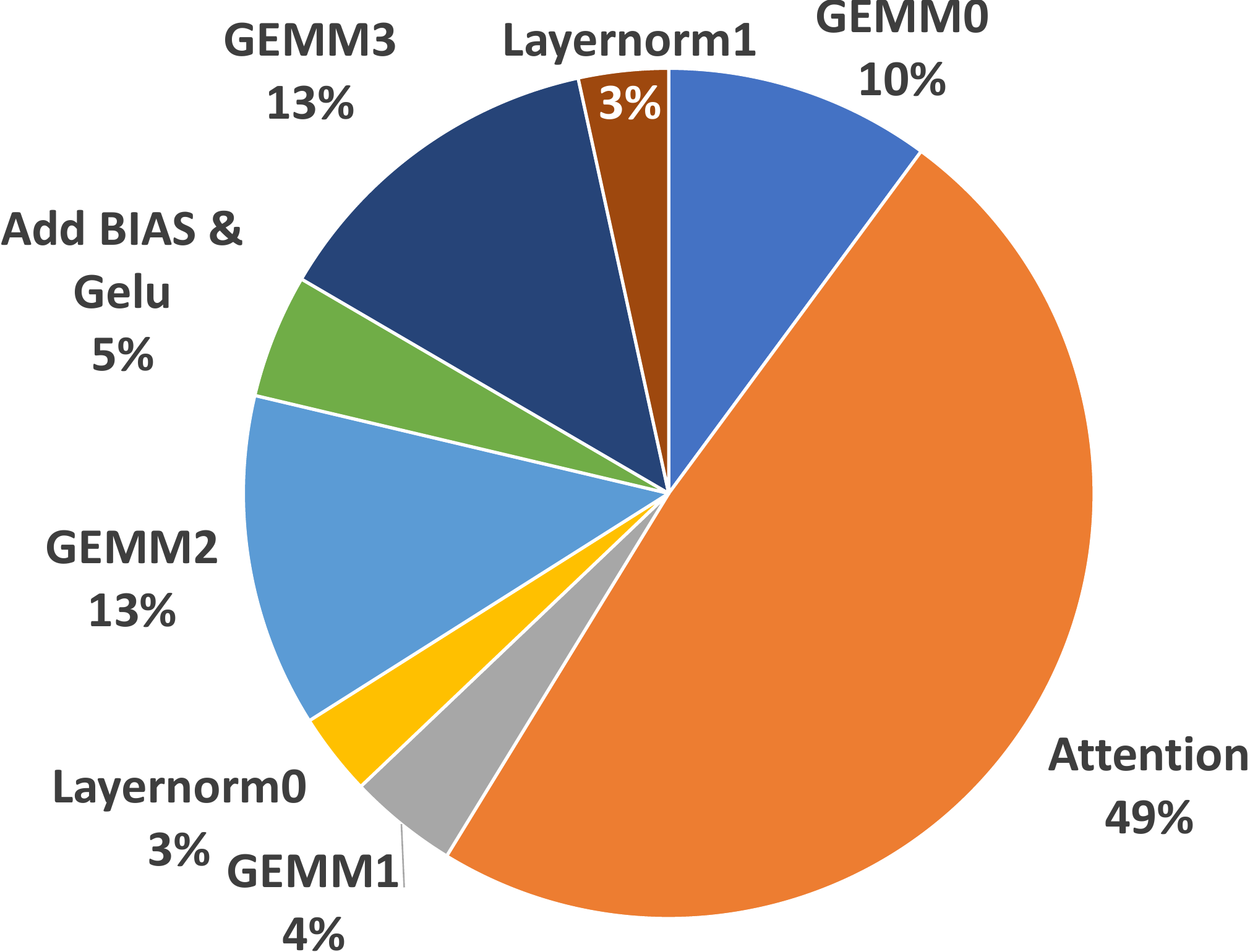}
}
\caption{Performance breakdown of forward BERT transformer.}
\label{fig:perf-breakdown}
\end{figure}

Figure \ref{fig:perf-breakdown} shows the performance breakdown for two sequence lengths. $\mathtt{GEMM0}$ to $\mathtt{GEMM3}$ refer to the consecutive four GEMMs that are enumerated from GEMM \#0 to GEMM \#3 in Figure \ref{fig:arch} (a). The other two batched GEMMs are part of the attention module and are therefore profiled together with the softmax as a whole, referred to as $\mathtt{MHA}$ in Figure \ref{fig:perf-breakdown}. The two sets of "add bias and layernorm" operations are referred to as $\mathtt{layernorm0}$ and $\mathtt{layernorm1}$. The profiling results show that the compute-bound GEMM operations account for 61\% and 40\% of the total execution time for both test cases. The attention module, which includes a softmax and two batched GEMMs, is the most time-consuming part of the transformer. As the sequence length increases to that of a GPT-2 model (1024), attention accounts for 49\% of the total execution time, while the remaining memory-bound operations (layernorm, add bias and activation) only take up 11\%-17\%.

\subsection{Fusing memory-bound operations of BERT transformer}

Since cuBLAS uses architectural-aware optimizations for high performance GEMMs, presumably there remain limited opportunities for further acceleration. Therefore, we turn our eyes to optimizing the modules containing memory-bound operations, such as attention (with softmax), feed forward network (with layernorm) and add bias followed by element-wise activation. We improve these operations by fusing distinct kernels and reusing data in registers to reduce global memory access. Figure \ref{fig:arch} (b) presents the BERT transformer pipeline with memory-bound kernel fusion, where we fuse layernorm and activation with their consecutive kernels.

\subsubsection{Add bias and layer normalization} These operations account for 10\% and 6\% of the overall execution time for sequence lengths 256 and 1024, respectively. After MHA, the result tensor ($\mathtt{valid\_word\_cnt \times hidden\_dim}$) needs to first be added upon the input tensor (bias) and perform layer normalization. Here hidden dimension ($\mathtt{hidden\_dim}$) equals $\mathtt{head\_num} \times \mathtt{head\_size}$. In standard BERT configuration, head number and head size are fixed to 12 and 64. The naive implementation introduces two rounds of memory access to load and store the tensor. We provide a fused kernel that only needs to access the global memory in one round to finish both layernorm and adding bias. Kernel fusion for this sub-kernel improves the performance by 61\%, which accordingly increases the single-layer BERT transformer performance by 3.2\% for sequence lengths ranging 128 to 1024 in average.

\subsubsection{add bias and activation} These operations account for 7\% and 5\% of the overall execution time for sequence lengths 256 and 1024, respectively. After the projection via matrix multiplication, the result tensor will be added against the input tensor and perform an element-wise activation using GELU \cite{hendrycks2016gaussian}. Our fused implementation, rather than storing the GEMM output to global memory and loading it again to conduct adding bias and activation, re-uses the GEMM result matrix at the register level by implementing a customized and fused CUTLASS \cite{nv-cutlass} epilogue. Experimental results validate that our fused GEMM perfectly hides the memory latency of bias and GELU into GEMM. After this step, we further improve the single-layer BERT transformer by 3.8\%.

\subsection{The zero padding algorithm for variable-length inputs}

\begin{figure}[ht] \centering
\includegraphics[width=0.48\textwidth]{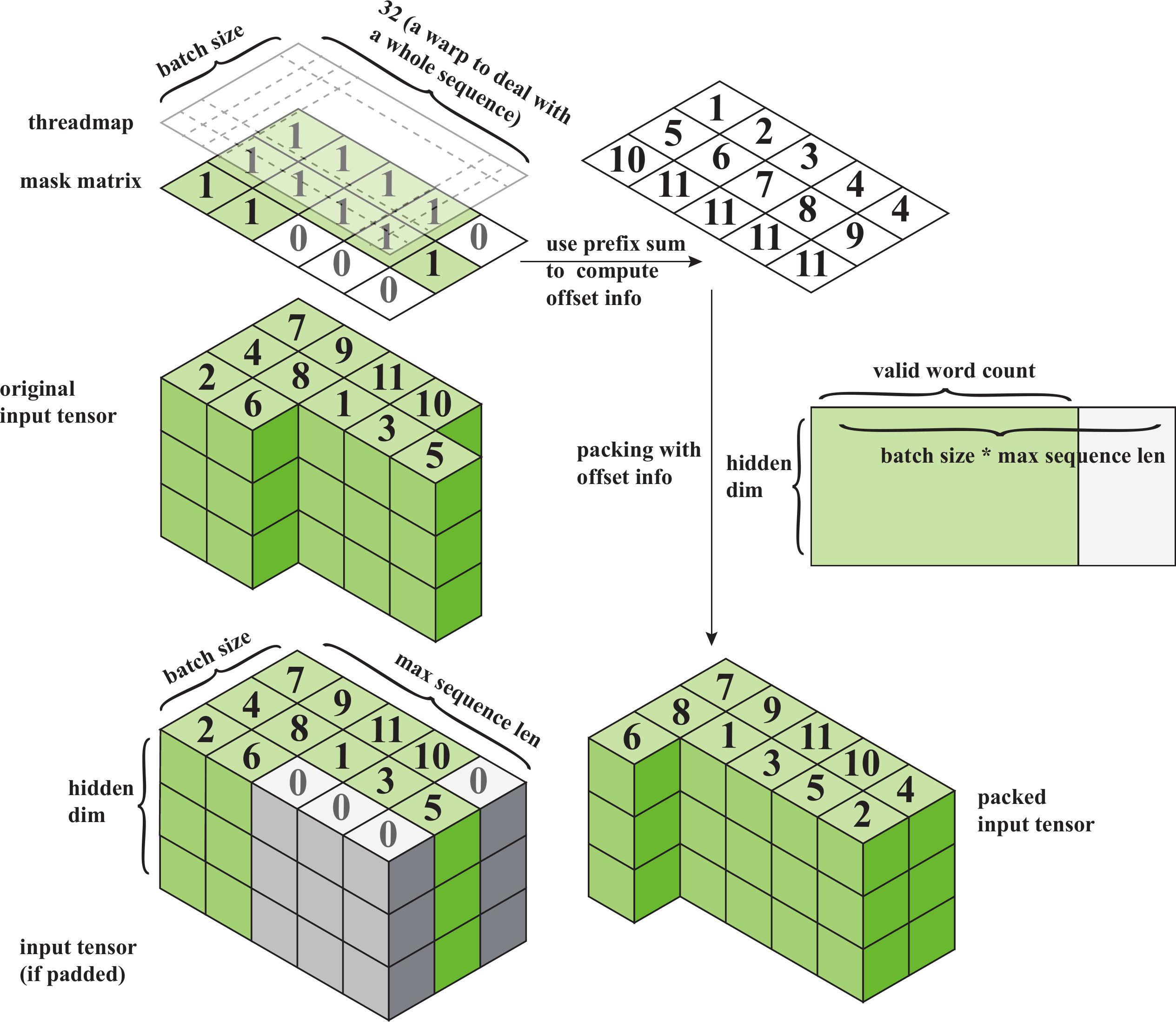}
\caption{The zero padding algorithm.}
\label{fig:remove}
\end{figure}

Because the real-time serving process receives sentences with various words as input tensor, the sequence lengths can often be different among batches. For such an input tensor composed of sentences with variable lengths, the conventional solution is to pad them to the maximal sequence length with useless tokens, which leads to significant computational and memory overhead. In order to address this issue, we propose the zero padding algorithm to pack the input tensor and store the positioning information for other transformer operations to index the original sequences.

Figure \ref{fig:remove} presents the details of the zero padding algorithm. We use an input tensor with 3 sentences (proceeded in 3 batches) as an example. The longest sentence contains 5 word tokens while the other two have 2 and 4 words. The height of the sample input tensor is 3, which is equal to the hidden dimension. The conventional method is to pad all sentences to the maximal sequence length by filling zeros. The elements, either 1 or 0, of the mask matrix correspond respectively to a valid token or a padded token of an input tensor with variable size. By calculating the prefix sum of the mask matrix, we can skip the padded tokens and provide the position indices of all valid tokens. We implement an efficient CUDA kernel to calculate the prefix sum and the position offset. Each warp computes the prefix sum for tokens of a whole sentence, so in total there are $\mathtt{batch\_size}$ warps assigned in each threadblock for prefix sum calculation. Once the prefix sum is computed, we pack the input tensor to a continuous memory area so that the total number of words used in future calculations is reduced from $\mathtt{seq\_len}\times\mathtt{batch\_size}$ to the actual valid word count of the packed tensor.

Figure \ref{fig:arch} (c) presents the detailed modifications on BERT by introducing our zero padding algorithm. Before conducting the positioning encoding, we calculate the prefix sum of the mask matrix to pack the input tensor so that we avoid computations on useless tokens in the first GEMM. Since batched GEMM in MHA requires identical problem shapes among different batches, we unpack the tensor before entering the attention module. Once MHA is completed, we pack the tensor again such that all remaining operations can benefit from the zero padding algorithm. The final result tensors are validated element-by-element against TensorFlow such that the correctness and accuracy are ensured. It is worth mentioning that padding and remove padding operations are fused with existing memory-bound footprints such as adding bias and transpose to minimize the overhead led by this feature.

Our presented padding-free algorithm is designed to ensure semantic preservation. We maintain an array that stores the mapping relationship of the valid tokens between the original tensor and the packed tensor. The transformer operates on the packed tensor, and intermediate operations, such as MHA, layernorm and activation, refer to this position array to ensure the correctness. At the end of each layer, we reconstruct the output tensor according to the position array such that the whole pipeline is semantic preserving.

\begin{table}[ht]
\centering
\begin{tabular}{|c|c|c|c|}
\hline
      & Baseline                             & Zero Padding        & Zero Padding + fused MHA \\ \hline
GEMM0 & $6mk^2$ & $6(\alpha\cdot m)k^2$ & $6(\alpha\cdot m)k^2$      \\ \hline
MHA   &    $4\frac{m^2}{bs}k$   &   $4\frac{m^2}{bs}k$   &  $4\frac{(\alpha\cdot m)^2}{bs}k$                \\ \hline
GEMM1 &   $2mk^2$    &   $2(\alpha\cdot m)k^2$     &   $2(\alpha\cdot m)k^2$       \\ \hline
GEMM2 &   $8mk^2 $           &    $8(\alpha\cdot m)k^2 $     & $8(\alpha\cdot m)k^2 $               \\ \hline
GEMM3 &    $8mk^2 $     &   $8(\alpha\cdot m)k^2 $     &   $8(\alpha\cdot m)k^2 $          \\ \hline
\end{tabular}
\caption{The computation number needed for variable-length inputs, where average sequence length = $\alpha$ * maximum, $m$ denotes $\mathtt{batch\_size}\cdot\mathtt{max\_seq\_len}$, $k$ is denote hidden dimension $\mathtt{head\_num}\cdot\mathtt{head\_size}$, $bs$ denotes the batch size.}
\label{tab:rm-padding-cnt}
\end{table}

Table \ref{tab:rm-padding-cnt} counts the floating point computations of a single-layer BERT transformer. The computations of memory-bound operations are not included since they are negligible compared with the listed modules. Enabling the zero padding algorithm eliminates redundant computations for all compute-intensive modules other than MHA due to the restrictions of batched GEMM. When the average sequence length is equal to 60\% of the maximum, turning on the zero padding algorithm further accelerates the BERT transformer by 24.7\%.

\subsection{Optimizing multi-head attention}

The zero-padding algorithm, although it effectively reduces wasted calculations for variable-length inputs, cannot directly benefit batched GEMM operations in MHA. This disadvantage becomes increasingly significant when the sequence length increases, as demonstrated in Table \ref{tab:rm-padding-cnt}. The complexity of MHA is quadratic to the sequence length, while the complexity of all other GEMMs is linear to the sequence length. This motivates us to provide a high-performance fused MHA while maintaining the benefits of the zero-padding algorithm. With our fused MHA, attention no longer faces redundant calculations on useless tokens, as shown in Table \ref{tab:rm-padding-cnt}.

\subsubsection{Unpadded fused MHA for short sequences}

For short input sequences, we hold the intermediate matrix in shared memory and registers throughout the MHA computation kernel to fully eliminate the quadratic memory overhead. We also access Q, K, and V tensors according to the positioning information obtained in the prefix sum calculation step to avoid redundant calculations on padding zeros for the MHA module.

\begin{algorithm}[ht]
    \caption{Unpadded fused MHA for short sequences}
    \label{alg:short-fmha}
    \DontPrintSemicolon
    \scriptsize
    \begingroup
    \color{teal}
    /* define skew offset to avoid bank conflict */ \\
    \endgroup 
    \texttt{\#define SKEW\_HALF 8} \\
    \textbf{\texttt{Shared memory}: }  \\
    \texttt{\_\_half s\_kv [max\_seq\_len][size\_per\_head + SKEW\_HALF];} \\
    \texttt{\_\_half s\_query [split\_seq\_len][size\_per\_head + SKEW\_HALF];} \\
    \texttt{\_\_half s\_logits [max\_seq\_len][size\_per\_head + SKEW\_HALF];} \\
    \begingroup
    \color{teal}
    /* warps collaboratively fill \texttt{s\_query} with adding bias fused */ \\
    \endgroup 
    \textcolor{blue}{Load} \texttt{\_\_half2 q\_bias} \\
    \For{seq\_id = warp\_id : warp\_num : split\_seq\_len}{
        \texttt{query = Q[batch\_seq\_offset + seq\_id + thread\_offset];} \\
        \texttt{offset = seq\_id*(head\_size+SKEW\_HALF)+(lane\_id*2);} \\
        \texttt{(\_\_half2 *)s\_query[offset] = fast\_add(query, k\_bias);} \\
    }
    \begingroup
    \color{teal}
    /* warps collaboratively fill \texttt{s\_kv} with adding bias fused */ \\
    \endgroup 
    \textcolor{blue}{Load} \texttt{\_\_half2 k\_bias} \\
    \For{seq\_id = warp\_id : warp\_num : batch\_seq\_len}{
        \texttt{key = K[batch\_seq\_offset + seq\_id + thread\_offset];} \\
        \texttt{offset = seq\_id*(head\_size+SKEW\_HALF)+(lane\_id*2);} \\
        \texttt{(\_\_half2 *)s\_kv[offset] = fast\_add(key, k\_bias);} \\
    }
    \begingroup
    \color{teal}
    /* compute Q*K using WMMA */ \\
    \endgroup
    Clear wmma fragment QK to zero \\
    \For{k\_id = 0 : head\_size / 16}{
        \textcolor{blue}{Load} \texttt{16x16 wmma fragments of Q} \\
        \textcolor{blue}{Load} \texttt{16x16 wmma fragments of K} \\
        Update QK = Q * K + QK using \texttt{wmma::mma\_sync}
    }
    Store fragment QK to \texttt{s\_logits} using \texttt{wmma::store\_matrix\_sync} \\
    \begingroup
    \color{teal}
    /* Compute softmax */ \\
    \endgroup
    \For{seq\_id = warp\_id : warp\_num : batch\_seq\_len}{
        \texttt{float logits[max\_seq\_len];} \\
        each thread loads a whole sequence to fill local registers \\
        \begingroup
        \color{teal}
        /* 1st round of reduction with register-level data re-use*/ \\
        \endgroup
        compute \texttt{max\_val} in local registers \\
        \begingroup
        \color{teal}
        /* register-level data re-use*/ \\
        \endgroup
        compute $P = exp(P - max)$ and update local registers \\
        \begingroup
        \color{teal}
        /* 2st round of reduction with register-level data re-use*/ \\
        \endgroup
        compute \texttt{sum\_val} in local registers \\
        \begingroup
        \color{teal}
        /* register-level data re-use*/ \\
        \endgroup
        compute $P = P / sum\_val$ and stream to \texttt{s\_logits} \\
    }
    \begingroup
    \color{teal}
    /* warps collaboratively fill \texttt{s\_kv} with adding bias fused */ \\
    \endgroup 
    \textcolor{blue}{Load} \texttt{\_\_half2 v\_bias} \\
    \For{seq\_id = warp\_id : warp\_num : batch\_seq\_len}{
        \texttt{value = V[batch\_seq\_offset + seq\_id + thread\_offset];} \\
        \texttt{offset = seq\_id*(head\_size+SKEW\_HALF)+(lane\_id*2);} \\
        \texttt{(\_\_half2 *)s\_kv[offset] = fast\_add(value, v\_bias);} \\
    }
    \begingroup
    \color{teal}
    /* Similar to Q * K so omitting the details here */ \\
    \endgroup 
    Compute P * V using \texttt{wmma} and stream to global memory
\end{algorithm}

\begin{figure}[ht] \centering
\scalebox{1.0}{
\includegraphics[width=0.48\textwidth]{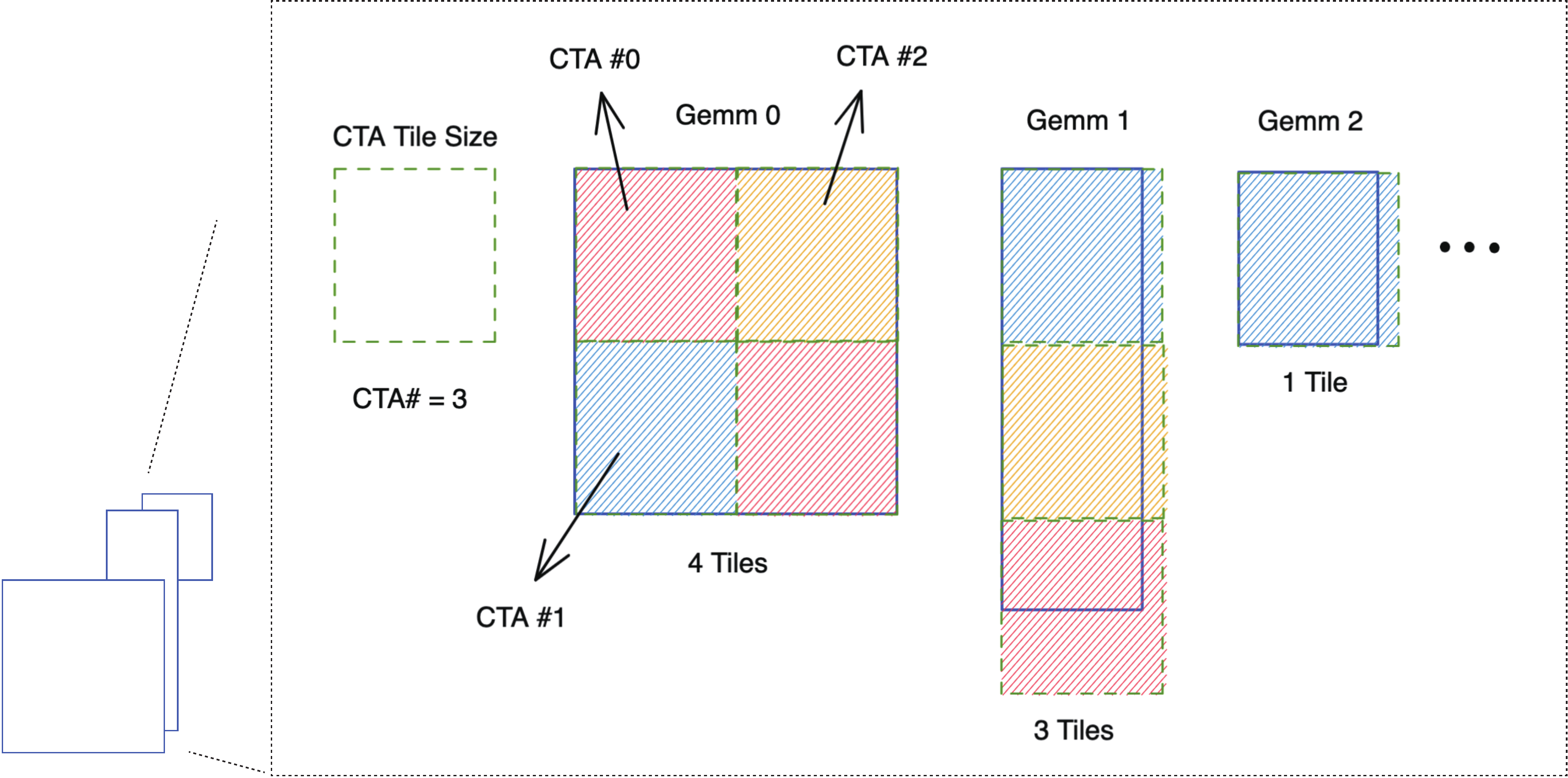}
}
\caption{Grouped GEMM demonstration.}
\label{fig:grouped-gemm}
\end{figure}

Algorithm \ref{alg:short-fmha} shows the pseudo code of our fused MHA for short sequences. We launch a 3-dimensional grid map: \{$\mathtt{head\_num}$,  $\mathtt{seq\_len}/\mathtt{split\_seq\_len}$, $\mathtt{batch\_size}$\}. Here $\mathtt{split\_seq\_len}$ is a user-defined parameter to determine the size of a sequence tile preceded by a threadblock (typically set to 32 or 48). The warp count of a threadblock is computed by the maximal sequence length: $\mathtt{split\_seq\_len} / 16 \times (\mathtt{seq\_len} / 16)$. Each threadblock loads a chunk of $Q$ ($\mathtt{split\_seq\_len}\times\mathtt{head\_size}$), $K$ ($\mathtt{max\_seq\_len}\times\mathtt{head\_size}$) and $V$ (($\mathtt{head\_size}\times\mathtt{max\_seq\_len}$)) into shared memory and computes MHA for a tile of the result tensor. We allocate three shared-memory buffers to hold $Q$, $K$, $V$ sub-matrices. Due to the algorithmic nature of MHA, we can re-use $K$ and $V$ chunks in the same shared-memory buffer $\mathtt{s\_kv}$. The intermediate matrix of MHA is held and re-used in another pre-allocated shared-memory buffer $\mathtt{s\_logits}$.

The workflow of fused MHA for short sequences is straightforward yet efficient. Each thread first loads its own tile of $Q$ and $K$ into shared memory and computes GEMM for $P = Q\times K$. The element-wise adding bias and scaling operations are both fused with the load process to hide the memory latency. GEMM is computed using the CUDA \texttt{wmma} intrinsic to leverage tensor cores of NVIDIA Ampere GPUs. The intermediate matrix $P$ is held in shared memory during the reduction. Because we explicitly design this algorithm for short sequences, each thread can load a whole sequence of $P$ from shared memory into register files for both reduction and element-wise exponential transform in softmax. Once the softmax operation is completed, we load a $K$ tile to shared memory to compute the second GEMM $O = P\times V$, and then store the result tensor $O$ to the global memory.

\subsubsection{Unpadded fused MHA for long sequences}

Because of the limited resources of register files and shared memory, the previous fused MHA is no longer feasible for long sequences. Therefore, we set 384 to be the cut-off sequence length and propose a grouped GEMM based fused MHA for large models.

\begin{figure}[ht] \centering
\includegraphics[width=0.48\textwidth]{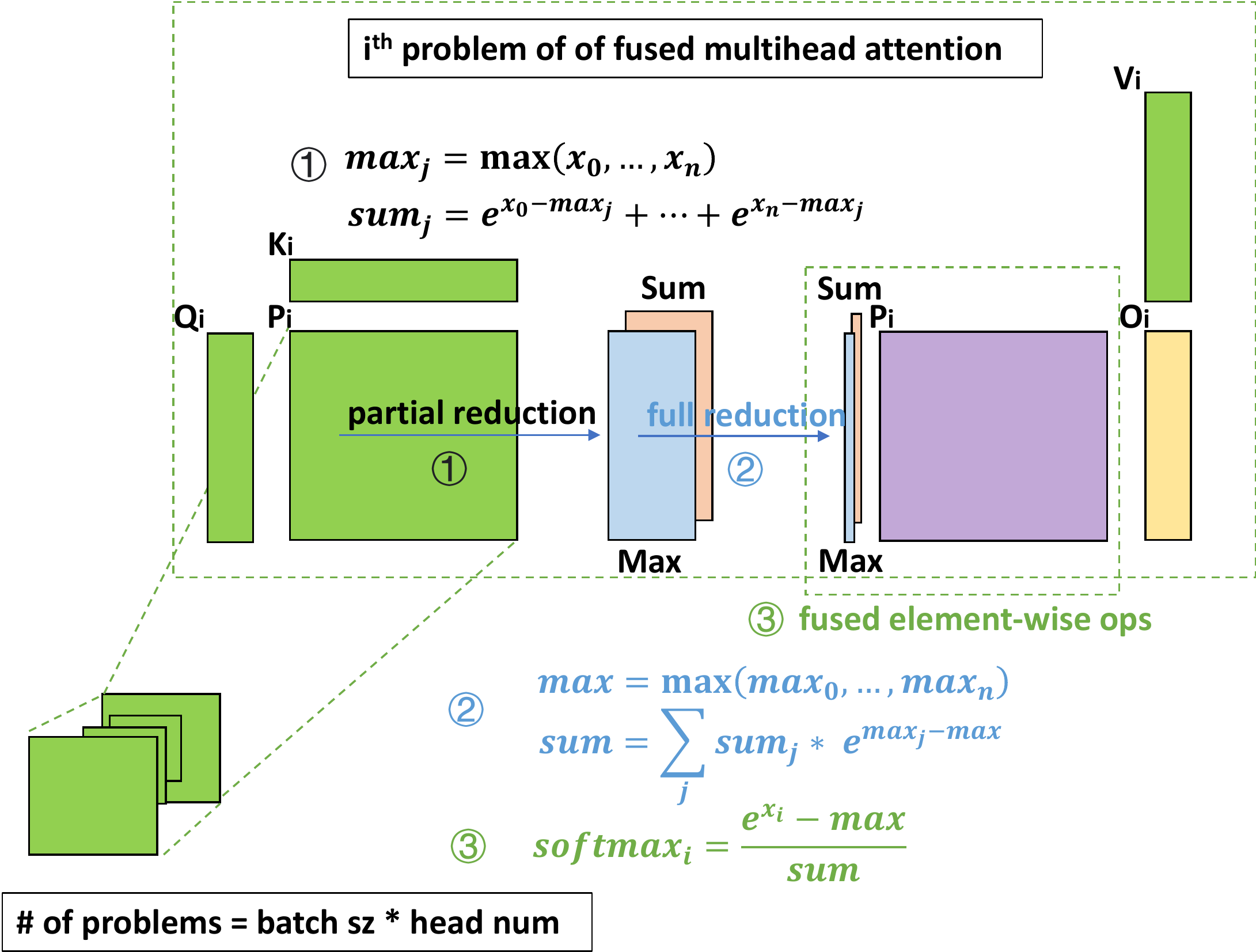}
\caption{Grouped-GEMM-based FMHA. The prototype of our fused MHA has been upstreamed to and released with CUTLASS 2.10. Source codes are available at \cite{nv-grouped-fmha}.}
\label{fig:grouped-fmha}
\end{figure}

The Grouped GEMM idea is first presented by NVIDIA CUTLASS \cite{nv-cutlass}. Different from batched GEMM, where all GEMM sub-problems are required to have an identical shape, grouped GEMM allows arbitrary shapes for sub-problems. This is enabled by a built-in scheduler that iterates over all GEMM sub-problems in a round-robin manner. Figure \ref{fig:grouped-gemm} demonstrates the idea of grouped GEMM using an example with 3 sub-problems. Supposing 3 threadblocks (CTAs) are launched, each CTA calculates a fix-sized CTA tile at each step until all GEMM sub-problems have been covered. GPU computes in waves, logically. In the first wave, All three CTAs calculate 3 tiles (light red, light yellow and light blue in the figure). And then in the second CTA wave, CTA \#0 moves to the bottom-right tile of GEMM 0 while CTA \#1 and CTA \#2 move to sub-problems of GEMM 1. In the final CTA wave, CTA \#0 and CTA \#1 continue to compute tasks in GEMM 1 and GEMM 2 while CTA \#2 keeps idle because there are no more available tiles in the computational graph.

Since grouped GEMM lifts the restriction on the shape of sub-problems, it can directly benefit MHA problems with variable-length inputs. Figure \ref{fig:grouped-fmha} presents our grouped-GEMM-based fused MHA for long sequences. The total number of MHA problems is equal to $\mathtt{batch\_size} \times \mathtt{head\_num}$. The MHA problems among different batches have different sequence lengths, while sequence lengths within the same batch are identical. The grouped GEMM scheduler iterates over all attention units in a round-robin manner. In each attention unit, we first compute GEMM $P_i = Q_i \times K_i$, and conduct softmax on $P_i$. The second GEMM $O_i = P_i \times V_i$ provides us with the final attention result. Here $i$ indicates the $i^{th}$ problem of grouped MHA with variable shapes. The softmax operation is fused with GEMMs to hide the memory latency. We have upstreamed the prototype of our grouped GEMM based fused MHA into NVIDIA CUTLASS \cite{nv-grouped-fmha}.

\begin{figure}[ht] \centering
\includegraphics[width=0.48\textwidth]{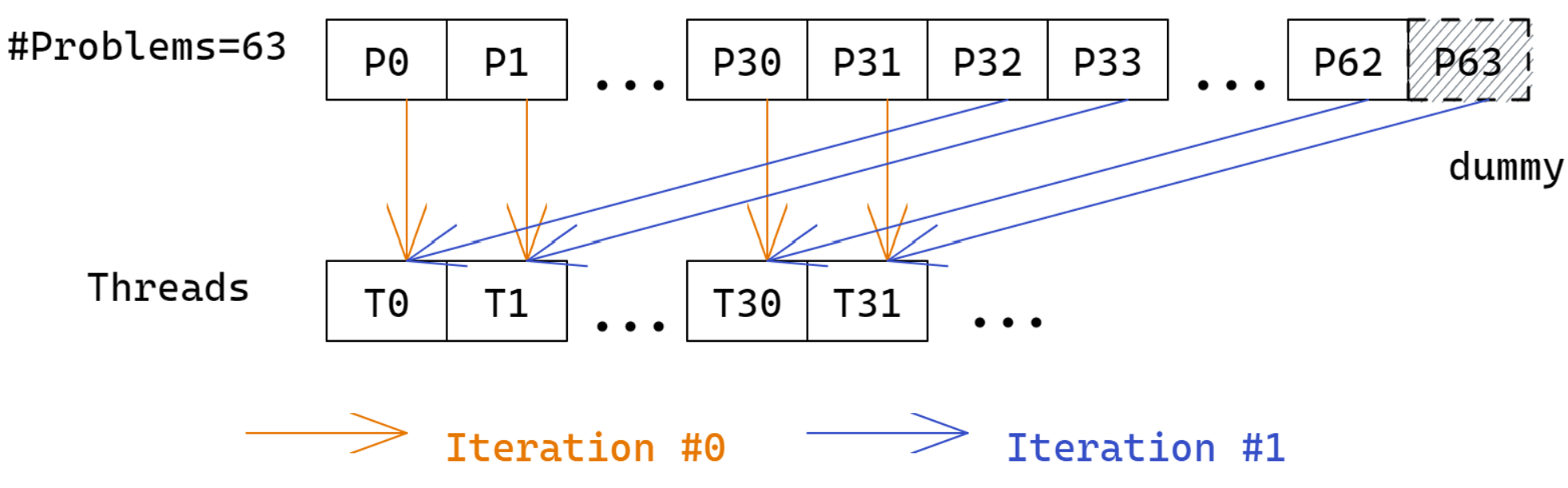}
\caption{Warp prefetching for grouped GEMM.}
\label{fig:problem-visitor}
\end{figure}

Grouped GEMM frequently checks with the built-in scheduler on the current task assignments, which leads to the runtime overhead. To address this issue, we propose an optimization over the built-in CUTLASS group GEMM scheduler. Figure \ref{fig:problem-visitor} shows our optimization for the original CUTLASS grouped GEMM scheduler. Rather than asking one thread to compute the current tasks metadata, we have all 32 threads in a warp compute the tile indices to visit at one time. Therefore, we achieve 32X fewer scheduler visit overhead. In practice, this strategy brings a $\sim$10\% improvement over the original CUTLASS grouped GEMM for standard BERT configurations. The prototype of this optimization has also been upstreamed to NVIDIA CUTLASS. We would refer an interested reader to \cite{nv-grouped-gemm} for detailed source codes.

\begin{figure}[ht] \centering
\includegraphics[width=0.48\textwidth]{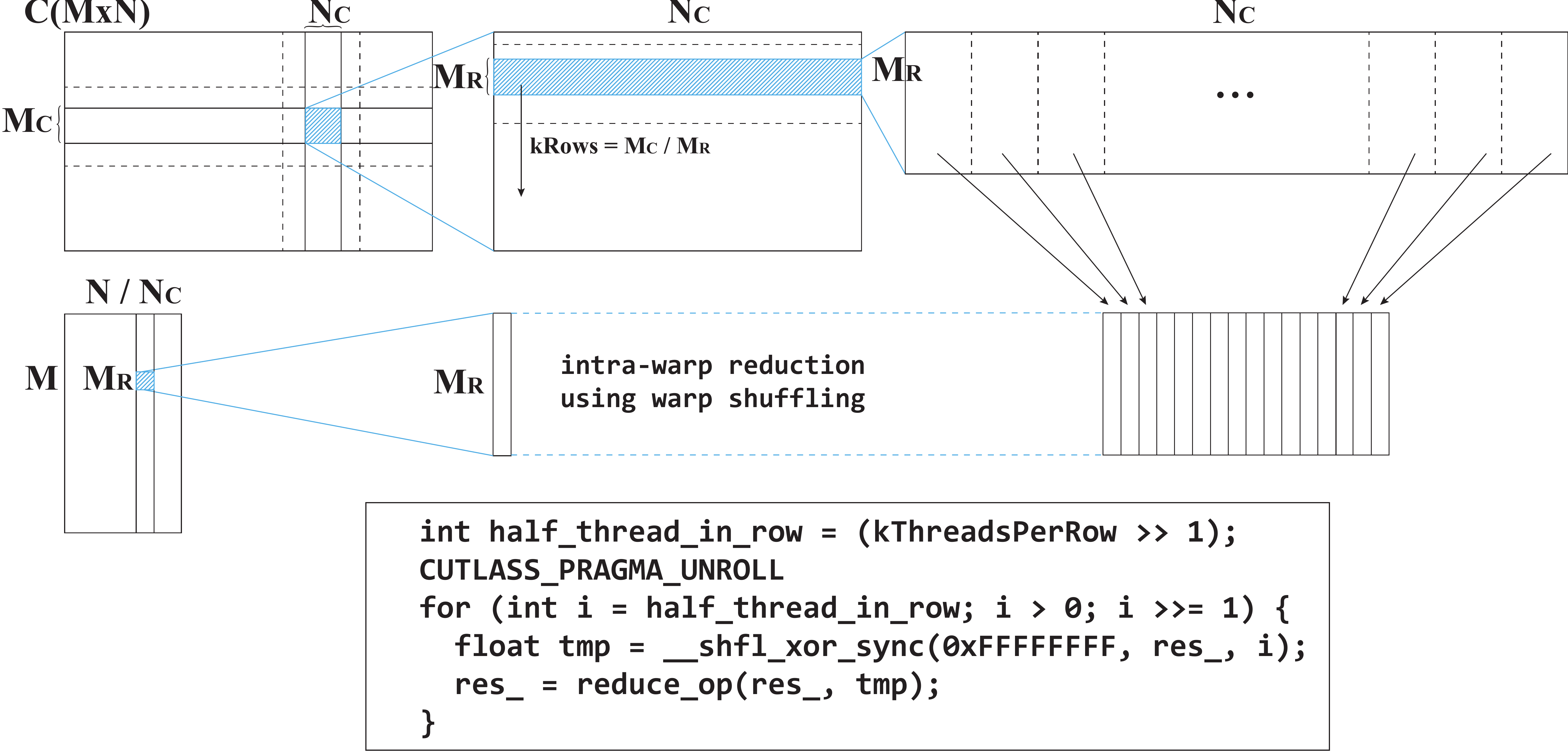}
\caption{Fused softmax reduction in grouped GEMM epilogue.}
\label{fig:epilogue-reduction}
\end{figure}

In addition to optimizing the grouped GEMM scheduler, we fuse the memory footprints of softmax into two grouped GEMMs of MHA. Figure \ref{fig:epilogue-reduction} shows the details of epilogue fusion for softmax reduction. A CTA computes an $M_C \times N_C$ sub-matrix. $M_C$ and $M_C$ are both set to $128$ to maximize the performance of GEMM. Under the default CUTLASS threadmap assignment, there are 128 threads per CTA, and the threadmap is arranged as $8\times16$, where each thread holds a 128-bit register tile in each step. After the intra-thread reduction, the $M_R\times N_C$ ($8\times128$) sub-matrix is reduced to $8\times16$, with one reduced result held by one thread. We then conduct an intra-warp reduction to further reduce from the column dimension, which is implemented via CUDA warp shuffling for efficiency. Similar reductions (intra-thread followed by intra-warp reduction) are performed to compute both max and sum in epilogue. Once max and sum are both reduced, we store them to global memory.

The reduction in epilogue only provides us with partial reduction within a threadblock because cross-threadblock communication is impractical under the current CUDA programming model. Hence, we need to launch a separated lightweight kernel, as shown in Figure \ref{fig:grouped-fmha}, to conduct the full reduction. In partial reduction, the target tensor of each attention unit is $\mathtt{seq\_len}\times\mathtt{seq\_len}$ while the full reduction just reduces a $\mathtt{seq\_len}\times\mathtt{seq\_len} / 128$. Therefore, the workload of full reduction is negligible to that of partial reduction In practice, the full reduction kernel only accounts for $\sim2\%$ of total execution time in fused MHA.


\begin{algorithm}[ht]
    \caption{Mainloop fusion of grouped FMHA}
    \label{alg:mainloop-fusion}
    \scriptsize
    \DontPrintSemicolon
    \textbf{\texttt{Register Tiles}: } \\ \texttt{WarpLoadedFragmentA warp\_loaded\_frag\_A[2]};\\
    \texttt{WarpLoadedFragmentB warp\_loaded\_frag\_B[2]};\\
    \texttt{\textcolor{red}{WarpLoadedFragmentNormSum  warp\_loaded\_frag\_norm\_sum}};\\
    \textbf{\texttt{Shared memory}: }  (kStages + 1) shared-memory tiles for A and B\\
    \begingroup
    \color{teal}
    /* prologue */ \\
    \endgroup 
    \textcolor{red}{Load k-invariant fused softmax tile to \texttt{warp\_loaded\_frag\_norm\_sum}} \\
    \textcolor{blue}{Prefetch} \texttt{kStages - 1} tiles of A to shared memory using \texttt{cp.async} \\
    \textcolor{blue}{Prefetch} \texttt{kStages - 1} tiles of B to shared memory using \texttt{cp.async} \\
    \textcolor{blue}{Prefetch} a tile of A from shared memory to \texttt{warp\_loaded\_frag\_A[0]} \\
    \textcolor{blue}{Prefetch} a tile of B from shared memory to \texttt{warp\_loaded\_frag\_B[0]} \\
    \begingroup
    \color{red}
    /* fused element-wise operation */ \\
    /* $A = \frac{exp(A - max)}{sum}$ */ \\
    \textcolor{red}{\texttt{elementwise\_transform(}} \\
    \textcolor{red}{\texttt{     warp\_loaded\_frag\_A[0],}} \\
    \textcolor{red}{\texttt{     warp\_loaded\_frag\_norm\_sum);}} \\
    \endgroup 
    \begingroup
    \color{teal}
    /* mainloop */ \\
    \endgroup 

    \For {$k$ \bf{to} \texttt{-kStages + 1}}
    {
        \begingroup
        \color{teal}
        /* Computes a warp-level GEMM */ \\
        /* with pipelined load during iterations */ \\
        \endgroup 
        \For {\texttt{warp\_mma\_k = 0} \bf{to} \texttt{kWarpGemmIterations - 1}}
        {
            \textcolor{blue}{Prefetch} \texttt{warp\_loaded\_frag\_A[\texttt{(warp\_mma\_k + 1) \% 2}]} \\
            \textcolor{blue}{Prefetch} \texttt{warp\_loaded\_frag\_B[\texttt{(warp\_mma\_k + 1) \% 2}]} \\
            \begingroup
            \color{red}
            /* fused element-wise transform */ \\
            \endgroup 
            \textcolor{red}{\texttt{elementwise\_transform(}} \\
                \textcolor{red}{\texttt{     warp\_loaded\_frag\_A[\texttt{(warp\_mma\_k + 1) \% 2}],}} \\
                \textcolor{red}{\texttt{     warp\_loaded\_frag\_norm\_sum);}} \\
            \begingroup
            \color{teal}
            /* Computes a warp-level GEMM*/ \\
            /* on data loaded in previous iteration */ \\
            \endgroup 
            \texttt{warp\_mma(}\\
            \texttt{    accum,}\\
            \texttt{    warp\_loaded\_frag\_A[\texttt{warp\_mma\_k \% 2}],}\\
            \texttt{    warp\_loaded\_frag\_B[\texttt{warp\_mma\_k \% 2}],}\\
            \texttt{    accum);}\\
            \textcolor{blue}{Prefetch} a tile of A to shared memory using \texttt{cp.async} \\
            \textcolor{blue}{Prefetch} a tile of B to shared memory using \texttt{cp.async} \\
        }
    }
\end{algorithm}

Once we have obtained the fully reduced $max$ and $sum$ vectors, we are ready to proceed element-wise transform $\frac{e^{x_{ij}} - max}{sum}$ on the first GEMM's output matrix. To hide the memory latency, we fuse these element-wise operations into the mainloop of the second GEMM. Algorithm \ref{alg:mainloop-fusion} presents our modifications (marked in red) of the original CUTLASS GEMM mainloop to enable softmax fusion. The original GEMM mainloop adopts the pipelining strategy to alleviate memory access latencies on both global memory and shared memory. For shared memory accesses, double register tiles are utilized to ensure that what is consumed in the current iteration has always been loaded in the previous iteration. For global memory accesses, a multi-stage loading strategy is employed with the help of the $\mathtt{cp.async}$ instruction of NVIDIA Ampere GPUs. The $\mathtt{cp.async}$ instruction allows loading data asynchronously from global memory to shared memory without consuming registers. Multiple such transactions can be proceeded concurrently, and a stage barrier ensures selected stages to be synchronized. The number of load stages ($\mathtt{kStages}$) is a compile-time constant defined by a user. Similar to shared memory accesses, loading from global memory is also pipelined to overlap memory latency with computation. Therefore, $\mathtt{kStages}$ pieces of shared memory buffers are needed under the multi-stage pipeline scheme. As shown in Algorithm \ref{alg:mainloop-fusion}, we preload the k-invariant vectors $sum$ and $max$ in prologue, and conduct element-wise transform right after the matrix elements are loaded into registers. Since the fused vectors are loaded outside of the GEMM mainloop, only negligible overhead is brought into the baseline GEMM and the memory latency to perform element-wise transform is perfectly hidden with GEMM computations.

The baseline MHA is a computational chain containing a batched GEMM, a softmax, and another batched GEMM. The time and memory complexity of all these operations are quadratic in the sequence length. Because the padding-free algorithm directly reduces the effective sequence length, MHA with variable-length input also gains a direct improvement. Our fused MHA, which is explicitly designed to handle both short and long sequences, incorporates the padding-free algorithm to alleviate the memory overhead of the intermediate matrix in MHA caused by padding for variable-length inputs. Our highly optimized MHA outperforms the standard PyTorch MHA by 6.13X and further accelerates the single-layer BERT transformer by 19\% compared to the previous step. As a result, this fully optimized version surpasses the baseline implementation in Figure \ref{fig:arch} (a) by 60\%. Since the remaining operations of a forward BERT transformer are all near-optimal GEMM operations, we conclude our optimizations at this step.

\section{Evaluation} \label{sec:results}

We evaluate our optimizations on an NVIDIA A100 GPU. The GPU device is connected to a node with four 32-core Intel Xeon Platinum 8336C CPUs, whose boost frequency is up to 4.00 GHz. The associated CPU main memory system has a capacity of 2TB at 3200 MHz. We compile programs using CUDA $\mathtt{11.6u2}$ with the optimization flag $\mathtt{O3}$. We compare the performance of ByteTransformer with latest versions of state-of-the-art transformers, such as TensorFlow $\mathtt{2.8}$, PyTorch $\mathtt{1.13}$, Tencent TurboTransformer $\mathtt{0.5.1}$, Microsoft DeepSpeed-Inference $\mathtt{0.7.7}$, and NVIDIA FasterTransformer $\mathtt{5.1}$. All the tensors benchmarked in this paper, unless specified, are in the half-precision floating-point format (FP16) to leverage tensor cores of NVIDIA GPUs. The variable sequence lengths in this section are generated randomly based on a uniform distribution with a range from 1 to the maximum length. We average the reported performance data over tens of runs to minimize fluctuations.

\subsection{Kernel fusion for layernorm and add-bias operations}

As depicted in Figure \ref{fig:arch}, BERT transformer is composed of a series of GEMM and memory-bound operations. Since GEMM are accelerated by near-optimal vendor's libraries cuBLAS and CUTLASS, we focus on optimizing the functional modules that involve memory-bound operations.

\begin{figure}[ht] \centering
\scalebox{1.0}{
\includegraphics[width=0.45\textwidth]{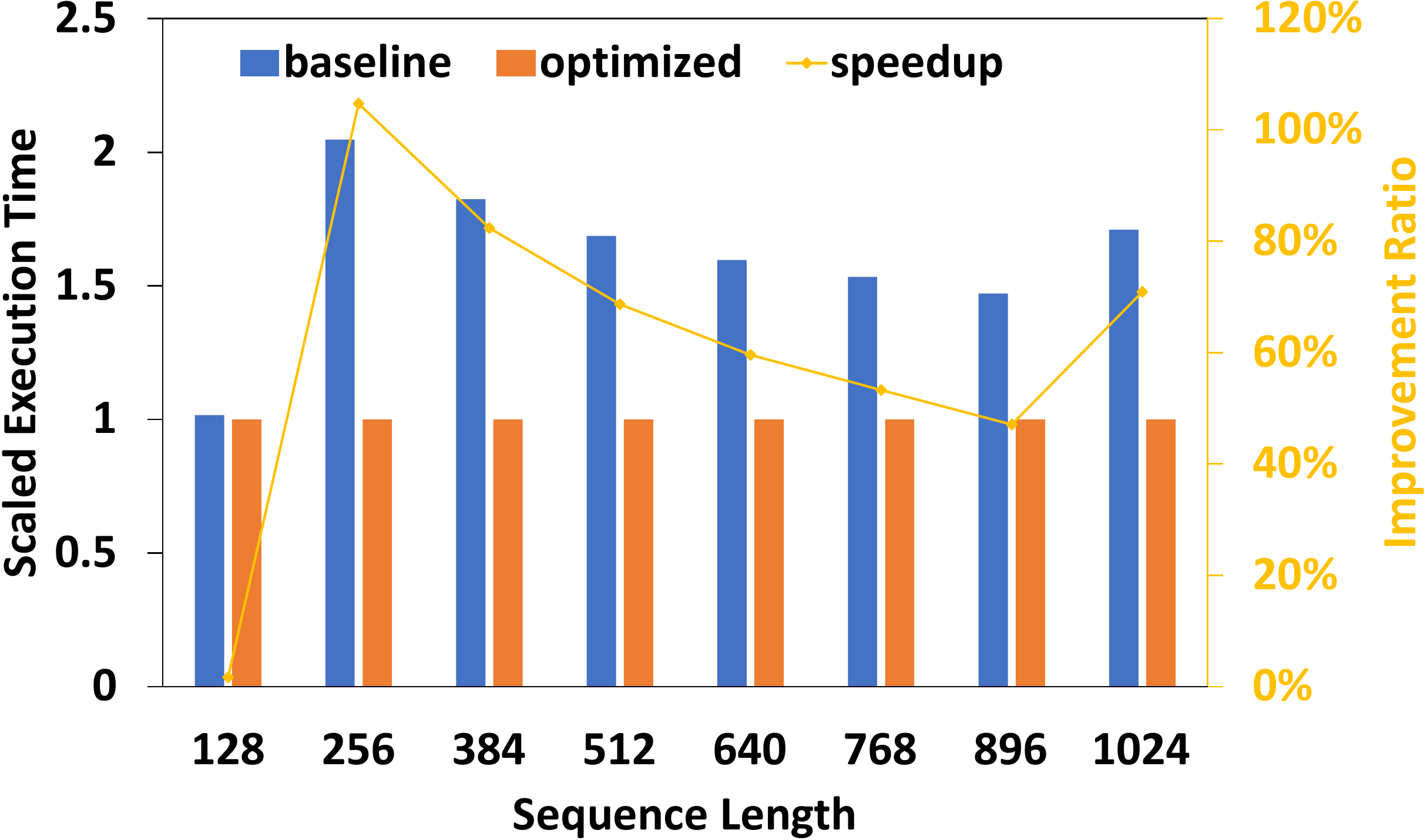}
}
\caption{Kernel fusion for add-bias and layernorm on a $(\mathtt{batch\_size}\cdot\mathtt{seq\_len})\times \mathtt{hidden\_dim}$ tensor. Here we profile for 16 batches with the hidden dimension fixed to 768 under the standard BERT configuration.}
\label{fig:layernorm}
\end{figure}

The result tensor needs to be added by the input tensor and normalized after projection and feed forward network of BERT transformer. Rather than launching two separated kernels, we fuse them into a single kernel and re-use data at the register level. In addition to kernel fusion, we leverage FP16 SIMD2 to increase the computational throughput of layernorm by assigning more workload to each thread. We normalize the execution time by that of the optimized layernorm and present the results in Figure \ref{fig:layernorm}: the improved version with kernel fusion provides us with a 69\% improvement on average over the unfused baseline for sequence lengths ranging 128 to 1024.

\subsection{Kernel fusion for GEMM and add-bias \& activation}

\begin{figure}[ht] \centering
\scalebox{1.0}{
\includegraphics[width=0.45\textwidth]{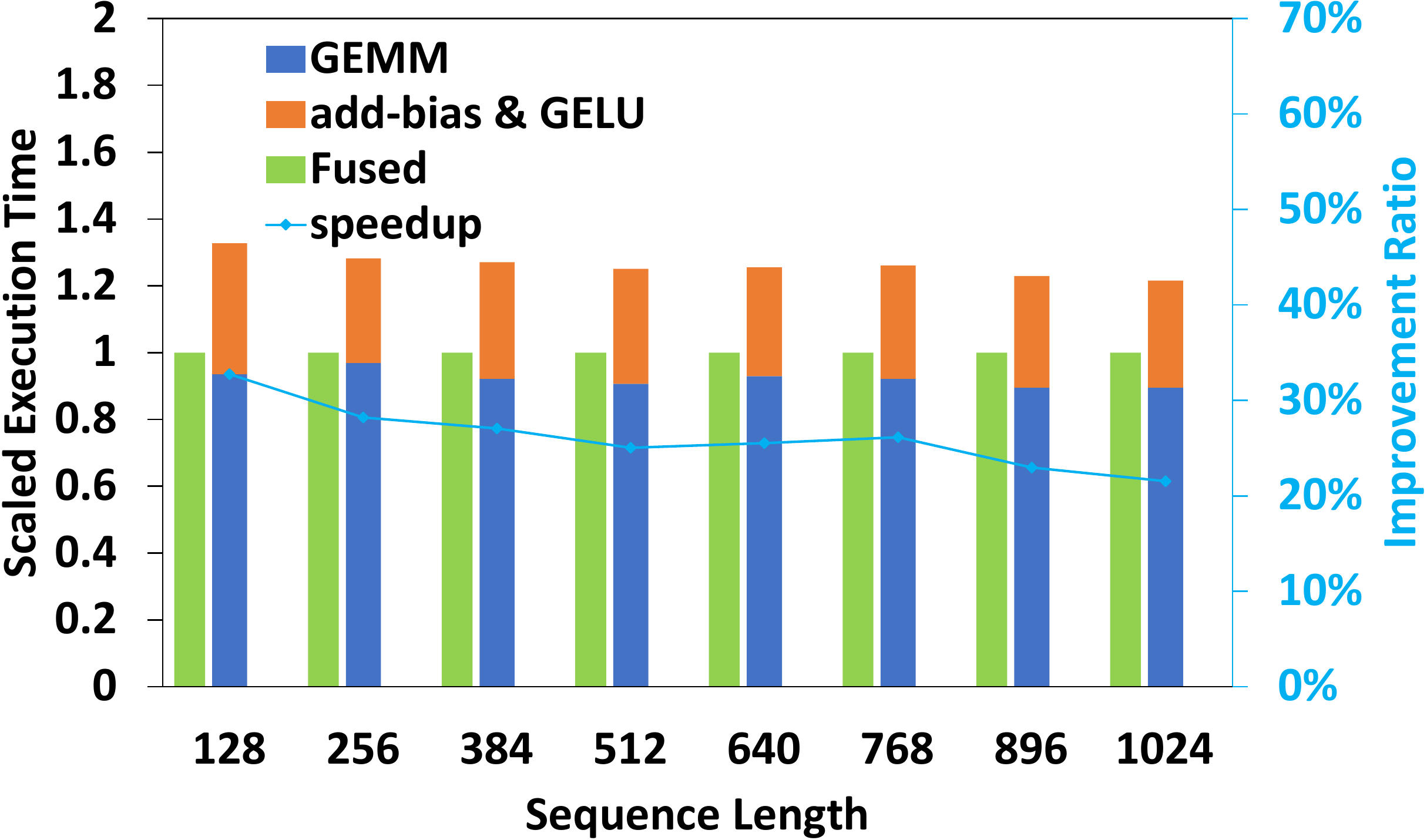}
}
\caption{Kernel fusion for GEMM, add-bias, and GELU. The shape of output tensor is $(\mathtt{batch\_size}\cdot\mathtt{seq\_len})\times (\mathtt{scale}\cdot\mathtt{hidden\_dim})$. Here we profile for 16 batches with the hidden dimension and the scale factor fixed to 768 and 4 under the standard BERT configuration.}
\label{fig:biasAct}
\end{figure}

Regarding the GEMM, add-bias and activation pattern in BERT transformer, we also provide a fused kernel to reduce the global memory access. An unfused implementation is to call vendor's GEMM, store the output to global memory, and then load the result matrix from global memory for further element-wise operations. In our optimized version, when the result matrix of GEMM is held in registers, we conduct fused element-wise operations that re-use data at the register level. Once the element-wise transform (add-bias and GELU) is completed, we then store the results to the global memory. Figure \ref{fig:biasAct} compares the performance of fused and unfused versions. In each clustered bar plot, the detailed execution time breakdown of the unfused implementation, normalized by the fused execution time (shown in the left bars), is shown in the stacked bar on the right. By fusing element-wise operations into the GEMM epilogue, we improve the performance by 24\% on average for sequence lengths ranging 128 to 1024. It is worth mentioning that we feed \textit{packed} tensors into both fused and non-fused kernels, such that the performance gain in Sec IV A and B are solely from kernel fusion.

\subsection{Optimizing multi-head attention}

Figure \ref{fig:perf-breakdown} shows that MHA accounts for 22\% - 49\% of the total execution time. We optimize this key algorithm by fusing softmax into GEMMs without calculating for useless padded tokens under variable-length inputs. For short sequences, we hold the intermediate matrix in registers and shared memory. For long sequences, we adopt a grouped GEMM based fused MHA and fuse softmax operations into our customized GEMM epilogue and mainloop to hide the memory latency. In both implementations, the input matrices are accessed according to the position information obtained from the zero padding algorithm so that no redundant calculations are introduced.

\begin{figure}[ht] \centering
\includegraphics[width=0.48\textwidth]{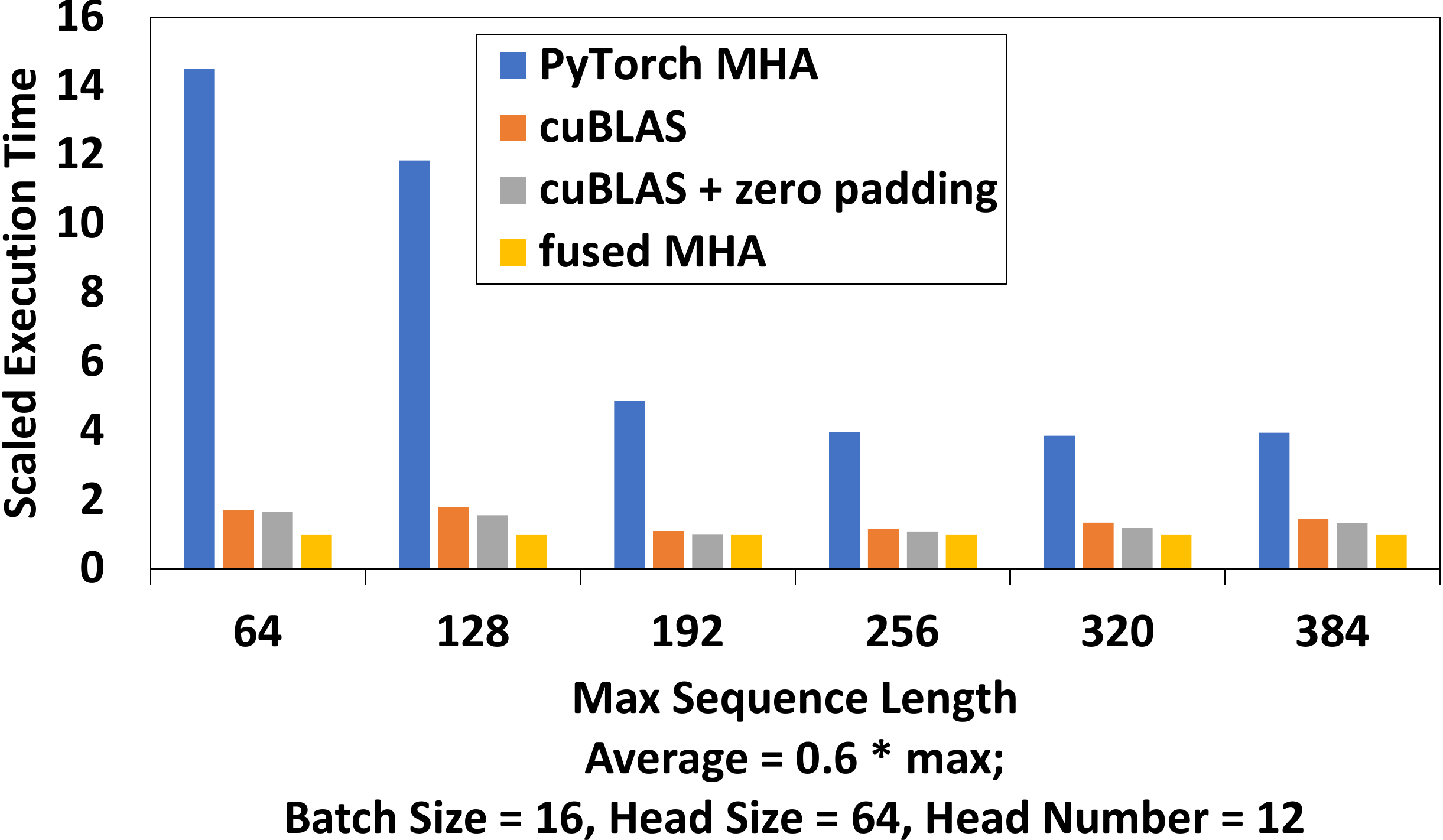}
\caption{Fused MHA for short sequences.}
\label{fig:fmha-short}
\end{figure}

Figure \ref{fig:fmha-short} compares the MHA performance for sequences shorter than 384. Here cuBLAS denotes the unfused implementation that calls cuBLAS for batched GEMM. The softmax operation between two batched GEMM can benefit from the zero padding algorithm, by only accessing unpadded tokens according to the known indices. This variant is denoted as \textit{cuBLAS + zero padding} in the figure. cuBLAS batched GEMM improves the performance over stand PyTorch MHA by 5 folds while enabling the zero padding algorithm for softmax further improves the performance by 9\%. Our MHA fully fuses the softmax and two batched GEMMs into one kernel, resulting in average speedups of 617\%, 42\%, and 30\% over all three variants for variable sequence lengths ranging from 64 to 384.

\begin{figure}[ht] \centering
\includegraphics[width=0.48\textwidth]{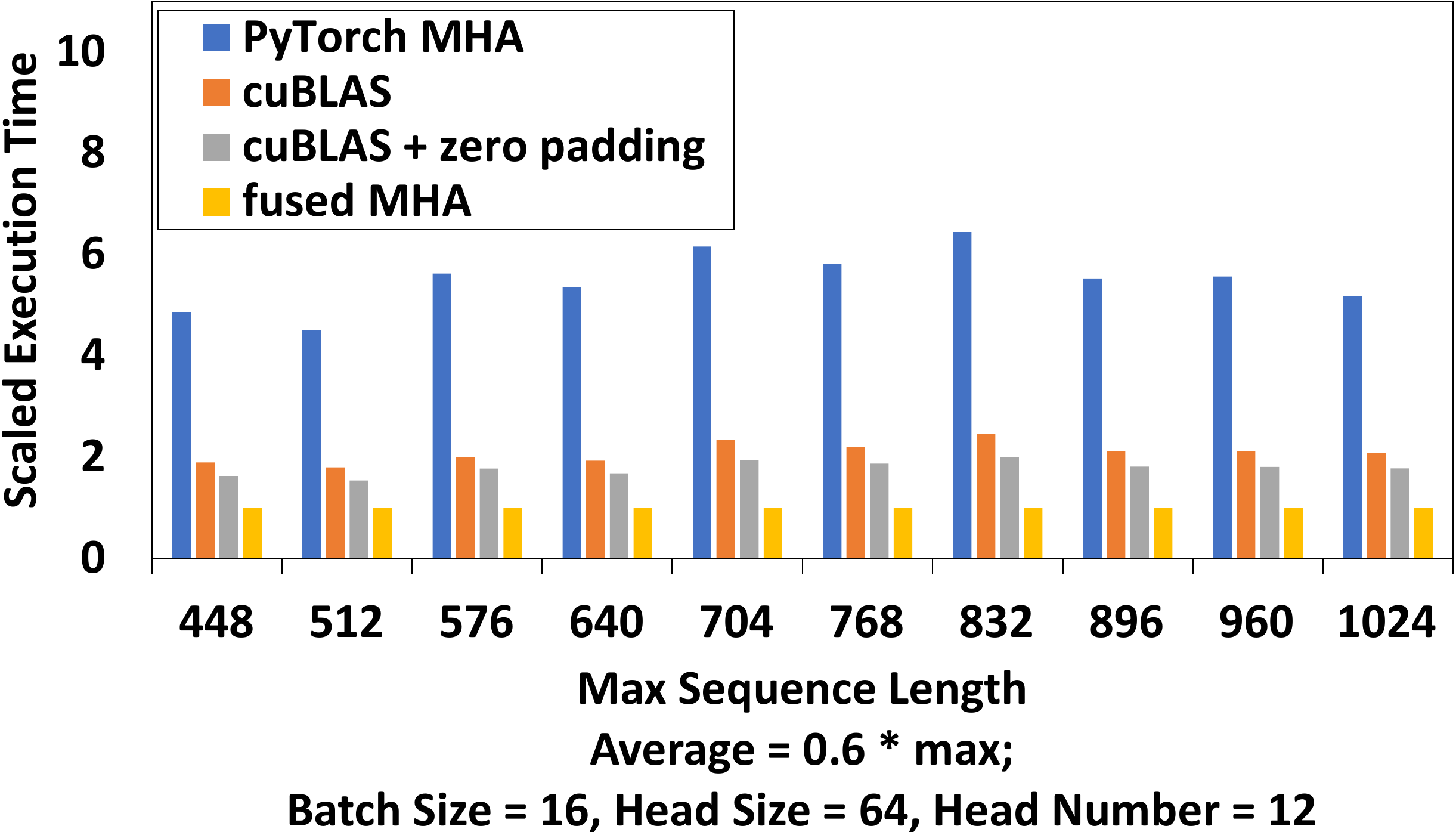}
\caption{Fused MHA for long sequences.}
\label{fig:fmha-long}
\end{figure}

Figure \ref{fig:fmha-long} compares the performance of the MHA for sequences longer than 448. The cuBLAS batched GEMM triples the MHA performance over PyTorch, while eliminating wasted calculations in softmax further brings a 17\% improvement. By introducing the high-performance grouped GEMM and fusing softmax into GEMMs, our fused MHA outperforms the variant MHA implementations by 451\%, 110\% and 79\% for maximal sequence lengths ranging 448 to 1024, where the average sequence length is 60\% of the maximum.

\begin{figure}[!ht] \centering
\includegraphics[width=0.48\textwidth]{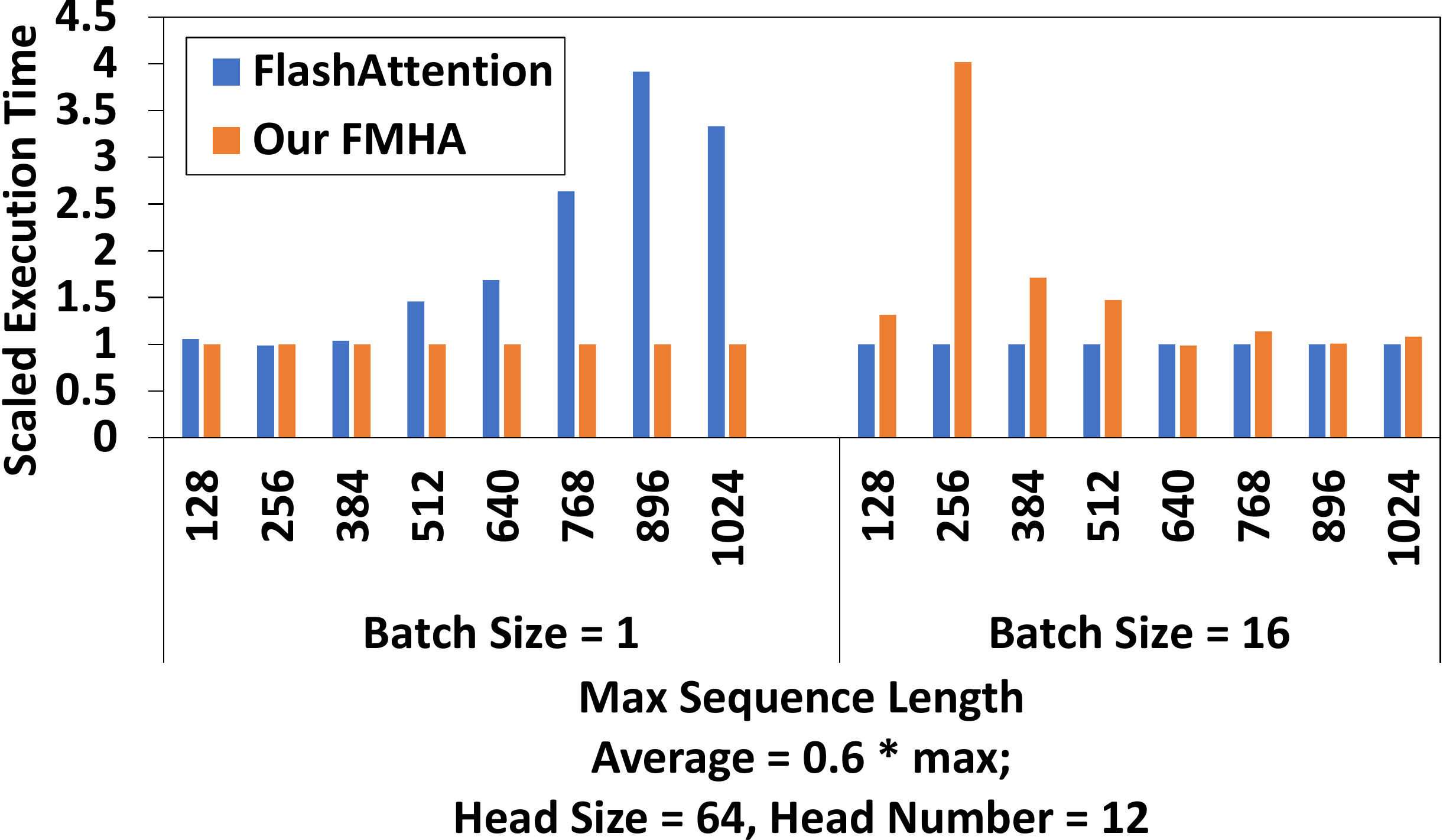}
\caption{Comparisons of our FMHA with FlashAttention.}
\label{fig:flash}
\end{figure}

Figure \ref{fig:flash} compares the scaled execution time of the FMHA module of our ByteTransformer against FlashAttention under the standard BERT setup. As shown in the figure, our FMHA presents advantages for small batch sizes (101\% faster on average) while FlashAttention becomes more efficient for large batch sizes (59\% faster on average). This is because FlashAttention maps a whole attention unit to a threadblock, which, although allows for the complete preservation of the intermediate matrix of an attention unit within shared-memory for any sequence length, results in performance degradation when there are insufficient tasks assigned.

\subsection{Benchmarking single-layer BERT transformer with step-wise optimizations}

\begin{figure}[ht] \centering
\includegraphics[width=0.48\textwidth]{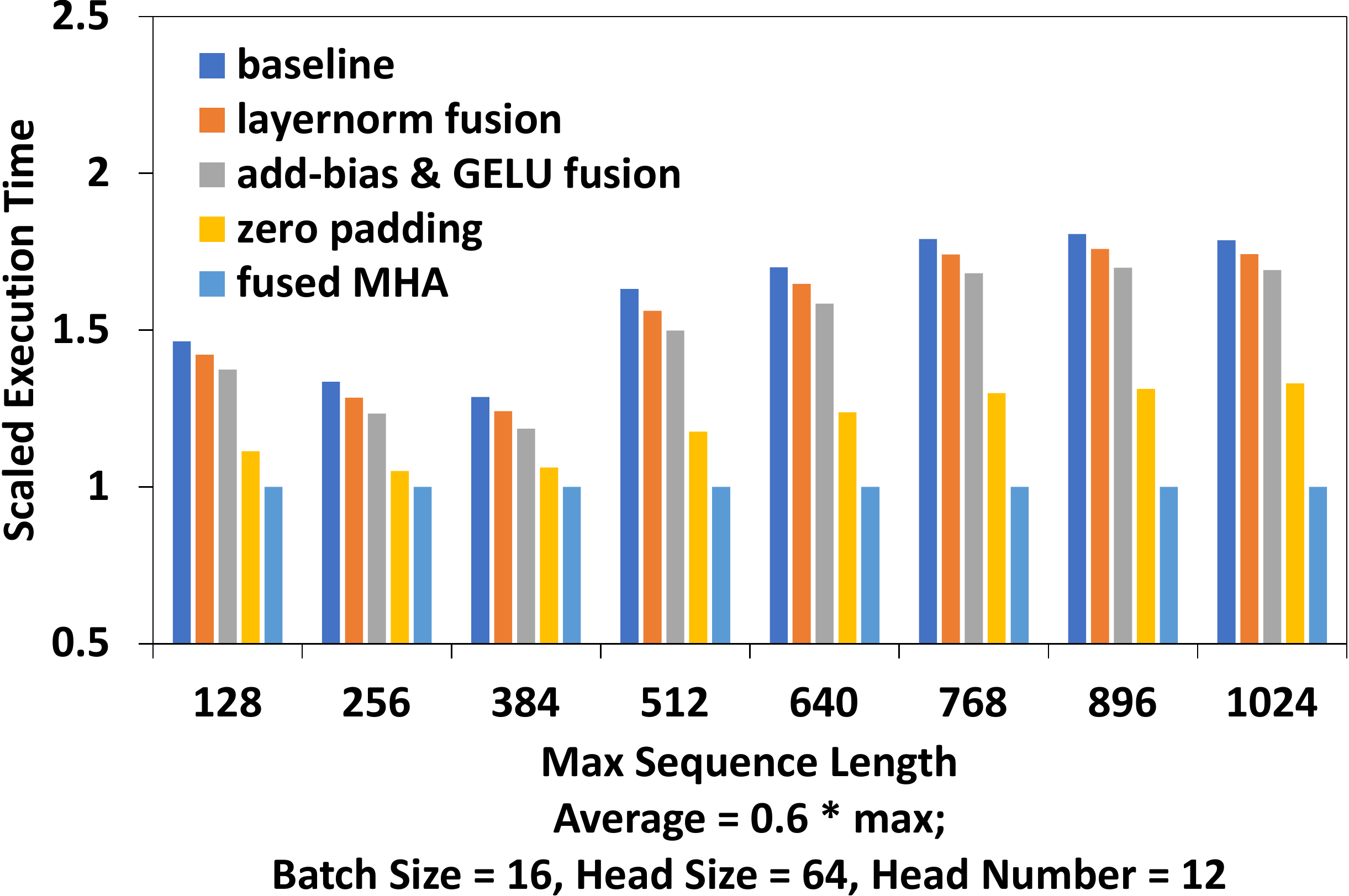}
\caption{Single-layer BERT transformer with step-wise optimizations. Each variant includes all previous optimizations.}
\label{fig:encoder}
\end{figure}

Figure \ref{fig:encoder} compares the performance of a single-layer BERT transformer to reflect our step-wise optimizations. At each step, we add a new optimization upon the previous variant. The baseline transformer implements the workflow in Figure \ref{fig:arch} (a) with padding. We then enable kernel fusion for adding bias and layernorm, which corresponds to \textit{layernorm fusion} in the figure. The next step is to fuse adding bias and GELU into GEMM, denoted by \textit{add bias \& GELU fusion}. In order to avoid calculating padded tokens for the variable-length inputs, we further propose the zero padding algorithm as shown in Figure \ref{fig:arch} (c). This is denoted by \textit{rm padding} in the figure. Our optimized transformer includes our high-performance fused MHA, as well as all previous optimizations.

Fusing adding bias and layernorm into one kernel improves the performance by 3.2\%. Fusing adding bias and activation into GEMM epilogue further improves the performance by 3.8\%. These two optimizations together improve the overall performance by 7.1\%. After bringing in the zero padding algorithm, the redundant calculations are eliminated in most modules other than MHA. We observe a 24\% improvement from the previous step. Finally, our fused MHA removes wasted calculations on padded tokens and enables an additional 20\% improvement. To summarize, the final version achieves 60\% improvement over the baseline version on single-layer BERT.

\begin{table}[ht] \centering
\caption{Single-layer BERT versus E.T. on A100.}
\begin{tabular}{l
    S[table-format=3] 
    S[table-format=3] 
    S[table-format=5] 
    }
\toprule
{Sequence Length}            & {E.T. (ms)}       & {ByteTransformer (ms)} & {Speedup}  \\ 
\midrule
{256} & {0.25}  & {0.07} & {3.57$\times$}                  \\ 
{1024} & {1.04} & {0.09} & {11.56$\times$}              \\ 
\bottomrule
\end{tabular}
\label{tab:et}
\end{table}

Table \ref{tab:et} compares the execution time for a single-layer, non-pruned BERT (batch size = 1) between E.T. and ByteTransformer, as E.T. has only open-sourced its single-layer, single-batch prototype. We achieve a speed-up of up to 11 times over E.T., which is optimized specifically for pruned models on legacy Volta GPUs. Since a pruned model can lead to significant reduction in total computations but with possible accuracy trade-offs, we do not include E.T. in our further end-to-end performance evaluations for non-pruned models on an A100 GPU for fairness and comparability.


\subsection{Benchmarking end-to-end performance of BERT}

\begin{figure}[ht] \centering
\subfigure[{Batch size = 1}]
{
\includegraphics[width=0.47\textwidth]{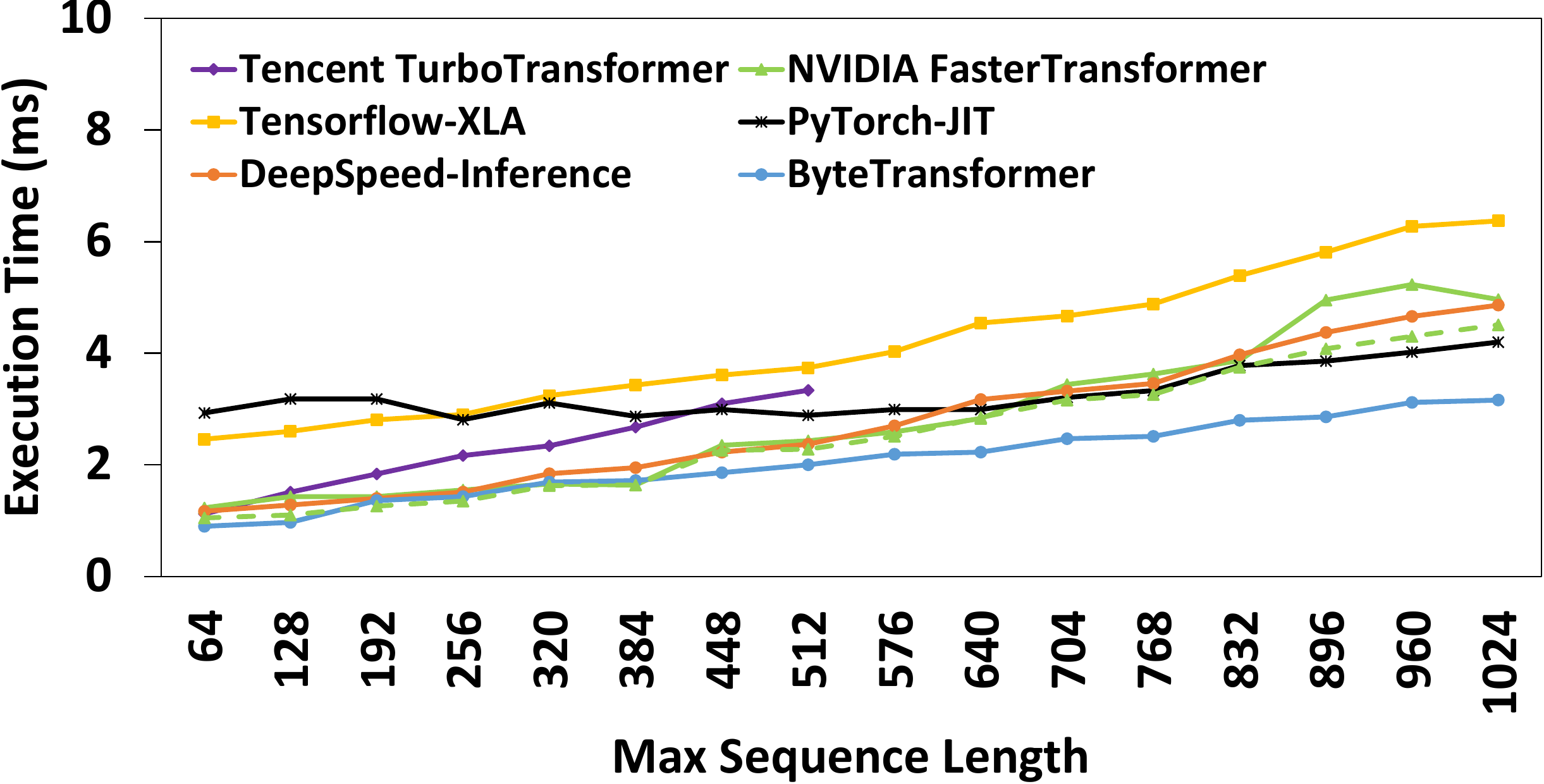}
}
\subfigure[Batch size = 8]
{
\includegraphics[width=0.47\textwidth]{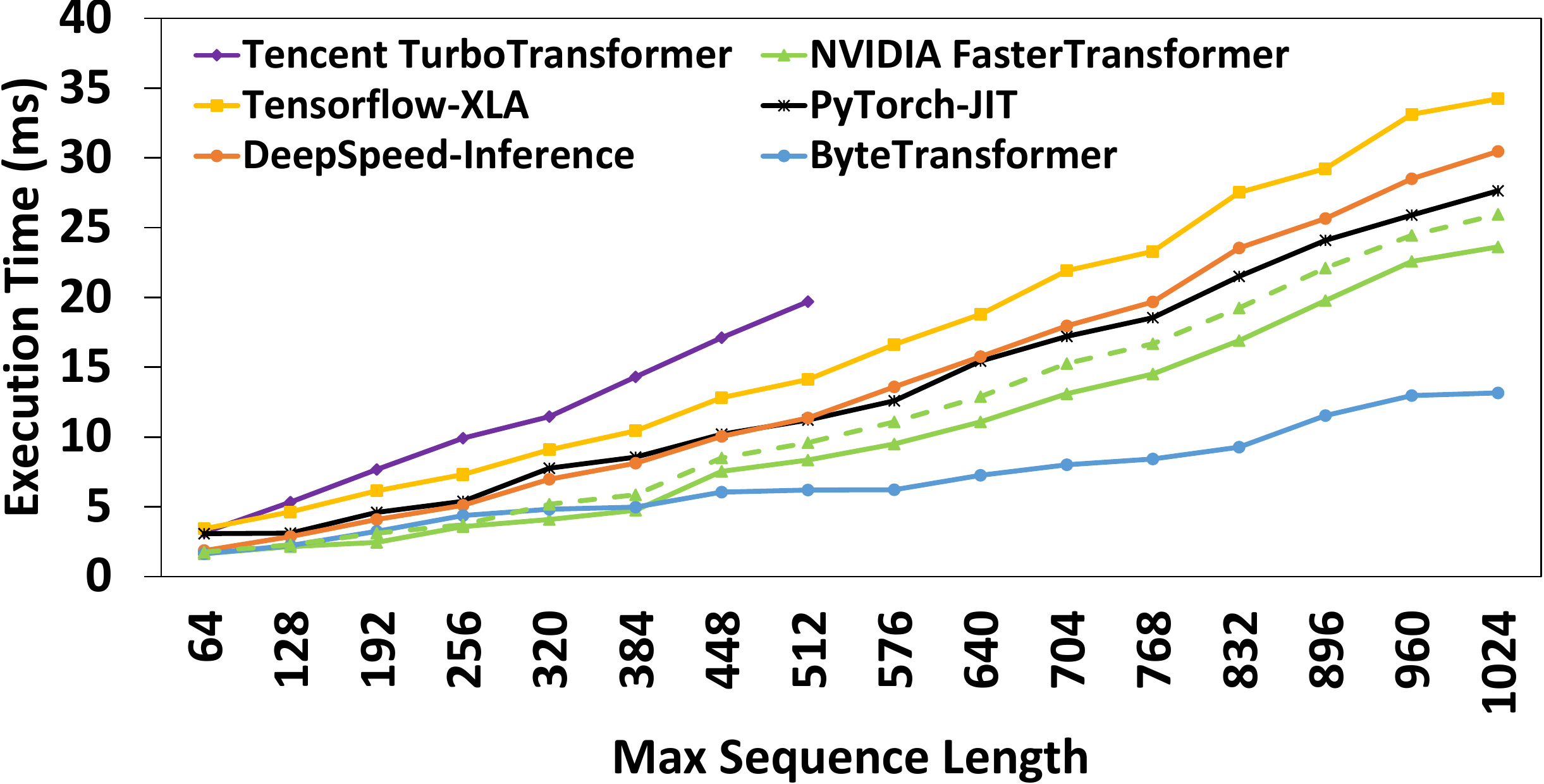}
}
\subfigure[Batch size = 16]
{
\includegraphics[width=0.47\textwidth]{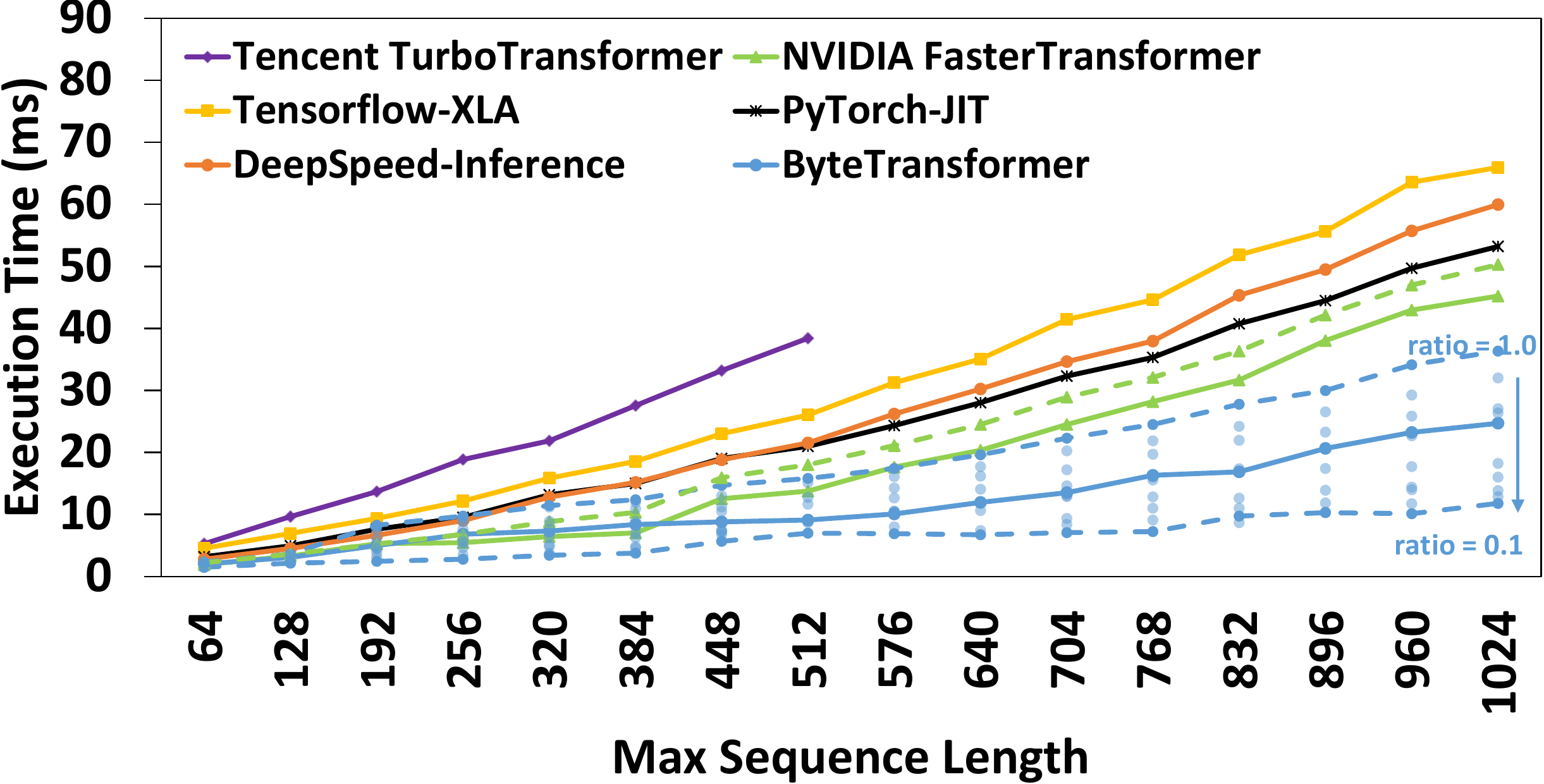}
}
\caption{End-to-end benchmark for standard BERT transformer, head size = 64, head number = 12, layer = 12, average sequence length = 0.6 * max sequence length.}
\label{fig:end-to-end}
\end{figure}

The standard BERT transformer is a stacked structure of 12 layers of the encoder module. The output of each encoder module is utilized as an input tensor in the next iteration. Figure \ref{fig:end-to-end} shows the end-to-end performance of ByteTransformer and compares it against state-of-the-art transformer implementations: PyTorch with JIT, TensorFlow with XLA acceleration, Micorsoft DeepSpeed-Inference, NVIDIA FasterTransformer and Tencent TurboTransformer. We adopt the standard BERT transformer configuration for end-to-end benchmark: 12 heads, head size equal to 64 and 12 iterations (layers). We benchmark for cases whose batch sizes are equal to 1, 8 and 16 and change sequence lengths from 64 to 1024.

Compared with popular DL frameworks PyTorch, TensorFlow, and Microsoft DeepSpeed-Inference, our ByteTransformer achieves 87\%, 131\%, and 74\% faster end-to-end performance on average. When benchmarking Tencent TurboTransformer, we turn on its $\mathtt{SmartBatch}$ mode to reach optimal batching performance. Since TurboTransformer only supports sequence lengths smaller than or equal to 512, we do not benchmark longer sequences for it. TurboTransformer re-groups and pads similar sequences into a batch so it launches excessive kernels at the run-time. It is faced with significant performance degradation for models with large batch numbers and sequence lengths. NVIDIA FasterTransformer, although it supports long sequences regarding the functionality, its back-end TensorRT fused MHA cannot be scaled to long sequences due to the limited register, its end-to-end efficiency cannot be maintained when the sequence length becomes longer than 512. Experimental results in Figure \ref{fig:end-to-end} show that ByteTransformer outperforms TurboTransformer and FasterTransformer by 138\% and 55\% on average, respectively.

Figure \ref{fig:end-to-end} (c) further includes the end-to-end performance of ByteTransformer for average-to-maximum sequence length ratios ranging from 0.1 to 1.0. The upper dashed blue line represents the execution time of ByteTransformer at a ratio of 1.0, while the lower dashed line corresponds to a ratio of 0.1. Our padding-free algorithm reduces the runtime by up to 66\% for a ratio of 0.1 compared to a fixed-sequence-length input. When disabling the support for variable-length inputs of FasterTransformer, as shown by the dashed green lines in Figure \ref{fig:end-to-end}, we observe a moderate decrease in performance for larger batch sizes (batch sizes = 8 and 16) but an improvement in performance for a small batch size (batch size = 1). In contrast, our FMHA-enabled padding-free algorithm significantly improves the performance of the end-to-end BERT transformer for variable-length input with an average-to-maximum ratio of 0.6, outpacing NVIDIA FasterTransformer by a notable difference of 54\% to 16\%.

\begin{table}[ht] \centering
\caption{Configurations of other BERT-like transformers.}
\begin{tabular}{l
    S[table-format=3] 
    S[table-format=3] 
    S[table-format=5] 
    }
\toprule
{Model}            & {layer number}       & {head number} & {head size}  \\ 
\midrule
{ALBERT} & {12}  & {16} & {64}                  \\ 
{DistilBERT} & {6} & {12} & {64}              \\ 
{DeBERTa} & {12}  & {12} & {64}              \\ 
\bottomrule
\end{tabular}
\label{tab:other-transformers}
\end{table}


\subsection{Extending to other BERT-like transformers}

\begin{figure}[ht] \centering
\includegraphics[width=0.48\textwidth]{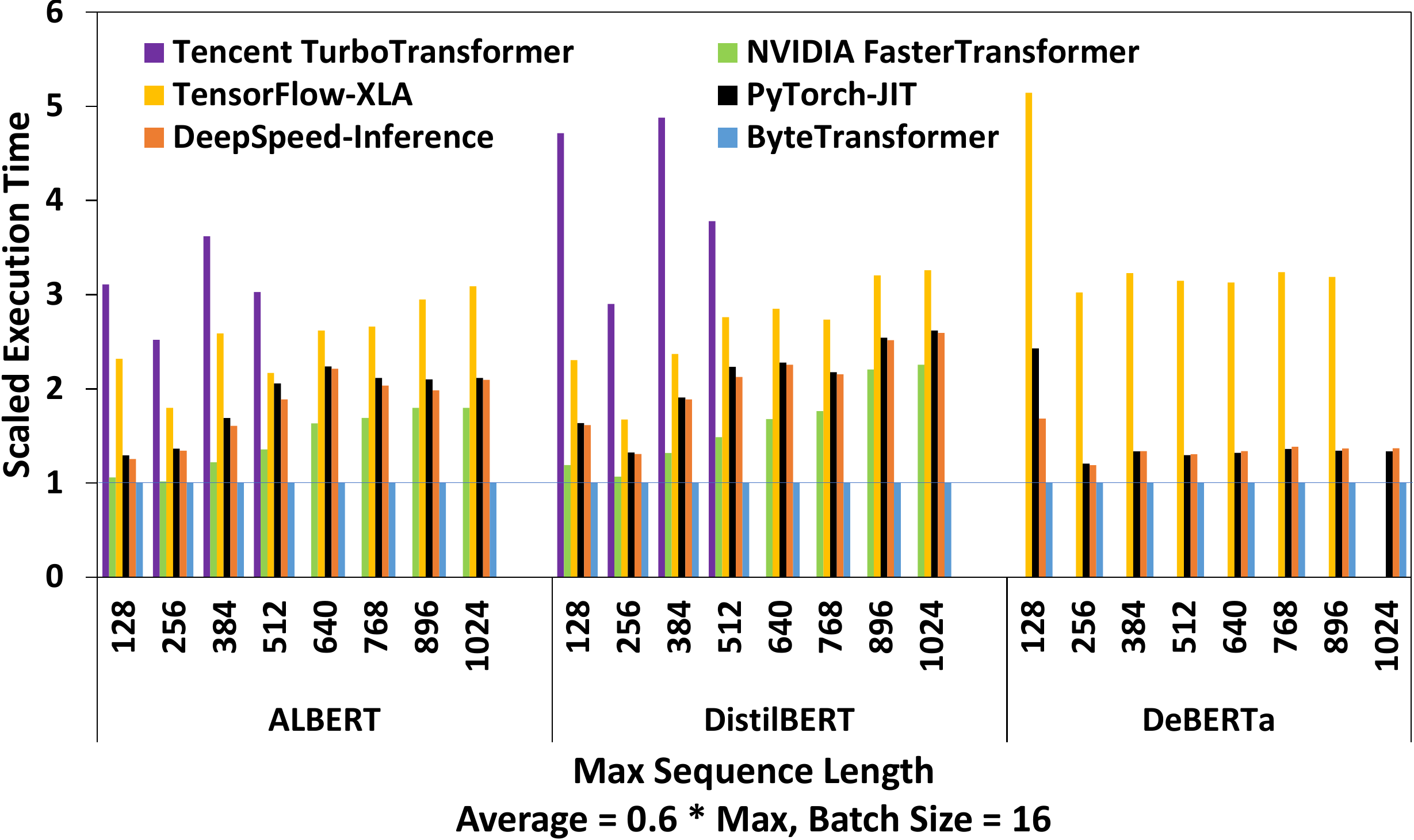}
\caption{End-to-end benchmark for other BERT-like models.}
\label{fig:other-bert}
\end{figure}

We extend the optimizations on kernel fusion and the padding-free algorithm presented in our work to other BERT-like transformers, including ALBERT, DistilBERT, and DeBERTa. Table \ref{tab:other-transformers} summarizes the model configurations, and readers can refer to \cite{lan2019albert, sanh2019distilbert, he2020deberta} for more detailed information about their architectures. Figure \ref{fig:other-bert} compares the performance of the ByteTransformer with state-of-the-art DL frameworks under these models. Following the setup for our demonstrated standard BERT benchmarks, the average sequence length is set to 60\% of the maximal sequence length. TurboTransfomer only supports sequences shorter than 512, so its performance data for long sequences are not presented. FasterTransformer and TurboTransformer do not support DeBERTa, so their results are not included in that model. It is worth noting that TensorFlow encountered an out-of-memory error for sequence length 1024 in the DeBERTa model, resulting in this data point being excluded. For ALBERT and DistilBERT, our ByteTransformer on average outperforms PyTorch, TensorFlow, Tencent TurboTransformer, DeepSpeed-Inference, and NVIDIA FasterTransformer by 98\%, 158\%, 256\%, 93\%, and 53\%, respectively. For the DeBERTa model, our ByteTransformer outperforms PyTorch, TensorFlow, and DeepSpeed by 44\%, 243\%, and 74\%, respectively.

\section{Conclusions} \label{sec:conclusion}
We have presented ByteTransformer, a high-performance transformer optimized for variable-length sequences. ByteTransformer not only brings algorithmic level innovation that frees the transformer from padding overhead, but also incorporates architecture-aware optimizations to accelerate functioning modules of the transformer. Our optimized fused MHA, as well as other step-wise optimizations, together provide us with significant speedup over current state-of-the-art transformers. The end-to-end performance of the standard BERT transformer benchmarked on an NVIDIA A100 GPU demonstrates that our ByteTransformer surpasses PyTorch, TensorFlow, Tencent TurboTransformer, Microsoft DeepSpeed-Inference, and NVIDIA FasterTransformer by 87\%, 131\%, 138\%, 74\% and 55\%, respectively. Moreover, we have shown that our optimizations are not specific to BERT, but can be applied to other BERT-like transformers, including ALBERT, DistilBERT, and DeBERTa. We are striving to make ByteTransformer completely open-source. This will allow the wider research community to benefit from our optimized implementation and to continue advancing the field. We are also dedicated to further expanding the presented strategies to accelerate a wider range of BERT-like transformer models, both in inference and training.




\bibliographystyle{./bibliography/IEEEtran}
\bibliography{./bibliography/example}

\end{document}